\documentclass[utf8]{FrontiersinHarvard}

\usepackage{url,hyperref,lineno,microtype}
\usepackage[onehalfspacing]{setspace}
\usepackage{xcolor, soul}
\usepackage{multirow}

\def\keyFont{\fontsize{8}{11}\helveticabold }
\def\firstAuthorLast{Hathaway {et~al.}} 
\def\Authors{Jamie Hathaway\,$^{1,2,\dagger}$, Abdelaziz Shaarawy\,$^{1,\dagger}$ , Cansu Akdeniz\,$^{1}$, Ali Aflakian\,$^{1,2}$, Rustam Stolkin\,$^{1,2}$, Alireza Rastegarpanah\,$^{1,2,*,}$ }

\def\Address{$^{1}$Department of Metallurgy \& Materials Science, University of Birmingham, Birmingham, B15 2TT, UK. \\
$^{2}$The Faraday Institution, Quad One, Harwell Science and Innovation Campus, Didcot, OX11 0RA, UK. \\
$^{\dagger}$These authors contributed equally to this work.
}

\def\corrAuthor{Alireza Rastegarpanah, Department of Metallurgy \& Materials Science, University of Birmingham, Birmingham, B15 2TT, UK. }

\def\corrEmail{a.rastegarpanah@bham.ac.uk}

\renewcommand{\vec}[1]{\boldsymbol{#1}}

\newcommand{\mat}[1]{\mathbf{#1}}
\newcommand{\tpose}{\mathsf{T}}

\begin{document}
\onecolumn
\firstpage{1}

\title[Tele-robotics for Battery Disassembly]{Towards Reuse and Recycling of Lithium-ion Batteries: Tele-robotics for Disassembly of Electric Vehicle Batteries}

\author[\firstAuthorLast ]{\Authors} 
\address{} 
\correspondance{} 

\extraAuth{}

\maketitle

\begin{abstract}

Disassembly of electric vehicle batteries is a critical stage in recovery, recycling and re-use of high-value battery materials, but is complicated by limited standardisation, design complexity, compounded by uncertainty and safety issues from varying end-of-life condition. Telerobotics presents an avenue for semi-autonomous robotic disassembly that addresses these challenges. However, it is suggested that quality and realism of the user's haptic interactions with the environment is important for precise, contact-rich and safety-critical tasks. To investigate this proposition, we demonstrate the disassembly of a Nissan Leaf 2011 module stack as a basis for a comparative study between a traditional asymmetric haptic-``cobot'' master-slave framework and identical master and slave cobots based on task completion time and success rate metrics. We demonstrate across a range of disassembly tasks a time reduction of 22\%--57\% is achieved using identical cobots, yet this improvement arises chiefly from an expanded workspace and 1:1 positional mapping, and suffers a 10--30\% reduction in first attempt success rate. For unbolting and grasping, the realism of force feedback was comparatively less important than directional information encoded in the interaction, however, 1:1 force mapping strengthened environmental tactile cues for vacuum pick-and-place and contact cutting tasks.

\tiny
 \keyFont{ \section{Keywords:} robotic disassembly, telerobotics, lithium-ion batteries, EV batteries, haptic, teleoperation}
\end{abstract}

\section{Introduction}

As a result of the increasing demand for electric vehicles (EVs) \citep{RIETMANN2020121038}, a large number of EV batteries are expected to reach end of life. Owing to a combination of contained high-value materials such as lithium, nickel and cobalt \citep{Thies2018}, and a limited lifespan of 10-15 years \citep{AI2019208}, there is an increasing research interest towards EV battery disposal. Battery disassembly is a critical step to enable gateway testing and sorting of end-of-life (EoL) battery components for re-use, and recovery of high-purity materials for recycling. This remains a predominantly manual process for trained personnel, requiring a high degree of precision and attention \cite{1401189, HybridDisassemblyEVBatteries}. However, EV batteries have numerous associated thermal and chemical hazards due to residual charge held within the battery and risk of thermal runaway. While previous studies such as \cite{LANDER2023120437} have emphasised the importance of autonomous disassembly for reducing disassembly cost, limitations are presented due to the high degree of variability in EV battery models \citep{Pehlken2017}, as well as the lack of dexterity alongside the highly dynamic, unstructured work environments---chiefly due to variety of models and manufacturers and lack of design standardisation. 
In consequence, hybrid frameworks for EV battery disassembly have been proposed in which robot and human work closely and collaboratively \citep{HybridDisassemblyEVBatteries}. However, the fact that robots perform tasks in the same environment as humans in production poses risks for humans. EV battery disassembly environments are considered hazardous for a human being to be present in; thus, direct collaboration with robots in such environments is critical. This is critical for damaged batteries which are challenging to disassemble due to further uncertainty in component end-of-life condition, which further compounds existing safety risks.

\textit{Tele-robotics} aims to mitigate the hazards of disassembly by enabling a human operator to carry out disassembly tasks remotely through a local interface while imparting some of the human operator's dexterity and fine motor control to the robot. Haptic devices are commonly employed in telemanipulation studies due to their ability to deliver force feedback to the operator while carrying out tasks. Nevertheless, one key challenge in haptic devices can be the master-slave asymmetry in regards of the kinematics \cite{9131861}; hence, there exists a range of mapping schemes by which the motion of the master device can be mapped to that of the slave. Previously, it has been emphasised the nature of this interface is important for carrying out dangerous/sensitive tasks \citep{BERNOLD2007518}. Furthermore, for disassembly in hazardous settings, it is suggested that the quality and realism of the feedback, and interface provided to the operator can greatly impact the level of task performance \citep{bolarinwa2022enhancing}. 

Using tele-robotics for the application of EV battery disassembly is not well explored, particularly, where damaged batteries present further critical challenges to the process of disassembly. Each stage of battery disassembly consists of a range of complex and precise contact-rich tasks which require a mixture of tactile and visual feedback to successfully complete disassembly. Given that not all EV battery disassembly tasks can be completed autonomously, this study demonstrates the application of tele-robotics to a range of battery disassembly tasks, such as unbolting, sorting and cutting. A series of tools were custom-designed in order to complete the tasks. 
We evaluate and compare the success rate and task execution time between a high-cost platform using two identical collaborative robots (``cobots'') and a relatively low-cost platform using a haptic device paired with a single cobot. We examine causative factors for differences in performance between these two platforms for each task. 

The remainder of the paper is structured accordingly: Section \ref{sec:RelatedWork} provides a survey of related studies in disassembly and telerobotics. The experimental setup for carrying out disassembly tasks is established in Section \ref{sec:ExperimentalSetup}, while Section \ref{sec:Methodology} introduces the control architecture for bilateral telemanipulation for both platforms. Section \ref{sec:Case Studies} defines the experimental case studies for different EV battery disassembly tasks, before evaluation of each telemanipulation platform on the basis of objective performance measures in Section \ref{sec:ResultsDiscussion}, and finally Section \ref{sec:Conclusion} concludes the paper. 

\section{Related Work} \label{sec:RelatedWork}

Methods of battery disassembly can be broadly categorised into fully-manual, fully-autonomous, and semi-autonomous approaches. Given the hazards that EV battery disassembly environments pose against human operators, the literature has been more oriented towards developing autonomous and semi-autonomous approaches on the basis of potential improvements in efficiency and safety. Autonomous approaches aim to increase the efficiency of disassembly by allowing robots to plan and carry out repetitive tasks in unstructured environments through the use of visual and tactile feedback. Examples of such works can be found in \cite{recycling7040048, Zhang2022AutonomousEV, met11030387, 9686083}. A common factor in these approaches is the use of labelling and detection methods to autonomously identify components and fasteners and construction of plans accordingly. However, this suffers from dependency on data and prior knowledge, and the risk of misidentification of battery components. For damaged batteries, existing prior knowledge datasets may not be suitable, or autonomous disassembly processes may fail; e.g. a fastener must be dislodged manually, or autonomous grasping fails due to structural deformation of components, requiring manual alignment. Furthermore, for sorting and lifting applications, the complexity of planning and collision avoidance remains an outstanding problem \cite{8055437,unknown}. Other approaches have aimed to reduce complexity of disassembly in various settings through tool design, such as gripper extensions \cite{8593567} or flexible grippers \cite{6088599}, to assist with unscrewing, grasping and other contact-rich manipulation tasks. However, even with such adaptations, autonomous approaches suffer from difficulties with generalisation to a range of tasks and across differing battery designs. Other studies such as \cite{batteries7040074} have examined optimisation of the EV battery disassembly process, emphasising the importance of battery design considerations for disassembly. As these are currently lacking for EVs, this introduces a high degree of variability and uncertainty into the disassembly process.

Recently, human-robot collaboration (HRC) has garnered attention for disassembly of end-of-life products, integrating both the robot's high efficiency in repetitive tasks and the human flexibility with higher cognition. However, there are few studies on this subject in the literature. In \citep{en15134856}, a high-speed rotary cutting wheel was adapted to perform robotic cutting at various points of the battery module casing. The proposed framework allowed a robot to efficiently carry out semi-destructive disassembly processes while allowing human operators to rapidly sort the battery components and remove connectors. Frameworks have further been proposed to semi-automate the process of extracting and sorting different objects from an EV battery pack using a mobile manipulator \cite{robotics10020082}. This study uses the behaviour tree model, which connects different robot capabilities, including navigation, object tracking and motion planning, for cognitive task execution and tracking in a modular format. Another study looked into human-/multi-robot- collaborations \cite{MultiSensorCobotDisassembly} in a disassembly scenario, where a technician and a number of robots coexist in the same environment interacting with each other to complete disassembly tasks. These presented articles demonstrate that HRC scenarios can offer both safety and time improvements over manual and autonomous disassembly strategies for EV LIBs. Nonetheless, issues such as coordination and allocation of tasks between robot and human and handling camera occlusion \cite{article} remain outstanding topics of research. Furthermore, in some cases, it is not possible for robots to be in the same environment as humans, as the environment is considered hazardous for humans.

Telemanipulation has been commonly employed in the nuclear industry for delicate handling of wastes that present a radiation hazard for human personnel. For example, the use of gloveboxes is examined in \cite{robotics10030085} in the context of telerobotics as a means of reducing radiation exposure risks to human personnel. Another study conducted real world experiments of a user performing nuclear decommissioning task via a unilateral teleoperated robotic system \cite{doi:10.1080/01691864.2023.2169588}. The study conducted a comparison analysis between three teleoperation scenarios: fully manual, teaching-based, and planning-based, in terms of safety, cognitive demand, and preparation time. Findings showed that planning-based is the most time efficient, yet lacks safety. Although these articles presented successful use cases of teleoperation in hazardous environments, bilateral systems were not investigated to show the importance of force feedback to the user on performance. Haptic devices have been extensively used in many studies developing bilateral teleoperation systems for various applications \cite{robotics10010029}. Haptic devices allow the operator to feel external force/torque, and further haptic cues that can be obtained in complex virtual environments. One study employed haptic devices as master devices to control dual 7-DOF serial arm manipulators to perform maintenance and repair tasks in nuclear power plants \cite{Ju2022}. They developed a shared bilateral teleoperation system including three elementary technologies: egocentric teleoperation, virtual fixture, and vibration suppression control to assist the human operator in performing shaft- and clutch-based peg-in-hole tasks. Furthermore, they conducted human-centric evaluations to measure the performance in terms of completion time, trajectory length, and human effort. The results of this study showed that haptic cues improved task performance significantly. Similar approaches have aimed to examine the effect of virtual constraints and force feedback guidance on task performance. Such approaches have been proposed in \cite{4483519}, based on imposition of rotational constraints on a component when axially aligned with its target. A related concept is proposed in \cite{7759628}, based on a shared control approach to remote object manipulation using visual information, and extended to a real environment in \citep{8594030}. In this way, the user is assisted during a grasping task by constraining the gripper orientation to the surface normal of a virtual sphere, centred on the object centre of mass. While performance improvements for these tasks have been documented, these approaches limit the flexibility provided to the operator; such constraints are often task-specific and susceptible to failure in edge cases, for example, if the autonomously identified grasp point is not suitable. Alternative approaches for sorting and separation of objects in an unstructured environment have been proposed based on haptic guidance cues \cite{8700204}. In this way, a visual inspection of the object is used to generate suitable grasp points, and the user is guided towards the suggested grasp point through tactile cues and vibration without constraining the free manipulation of the robot.

The majority of considered studies have focused on telerobotics for assembly tasks, nuclear decommissioning and waste handling applications. However, there is a paucity of research surrounding the use of telerobotics in an EV disassembly context. Furthermore, while many studies have explored the effect of shared autonomy and control system design on task performance, a limited number of works have investigated the nature of the interface exposed to the operator and its effects on task performance. Such a comparison has been explored in \cite{ComparisonTeleopKeyboardVsHaptic} between a keyboard-mouse setup, XBOX controller and a bilateral setup using a haptic device. For a range of patient care tasks, involving grasping and moving a remote and adjusting a camera view, a haptic device was found to improve the speed at which operators carried out tasks, however no subjective differences in mental load were identified. In \cite{8979376}, a framework was presented using a pair of identical cobots, and explores human-centric performance metrics between methods of rendering guidance and environmental forces to the user on the performance of a peg-in-hole task. However, the range of tasks remains limited and the effects of the interface remain unclear for the wider range of tasks featuring in a disassembly environment.

\section{Experimental Setup} \label{sec:ExperimentalSetup}

\begin{figure}[h]
    \centering
    \includegraphics[width=1\textwidth]{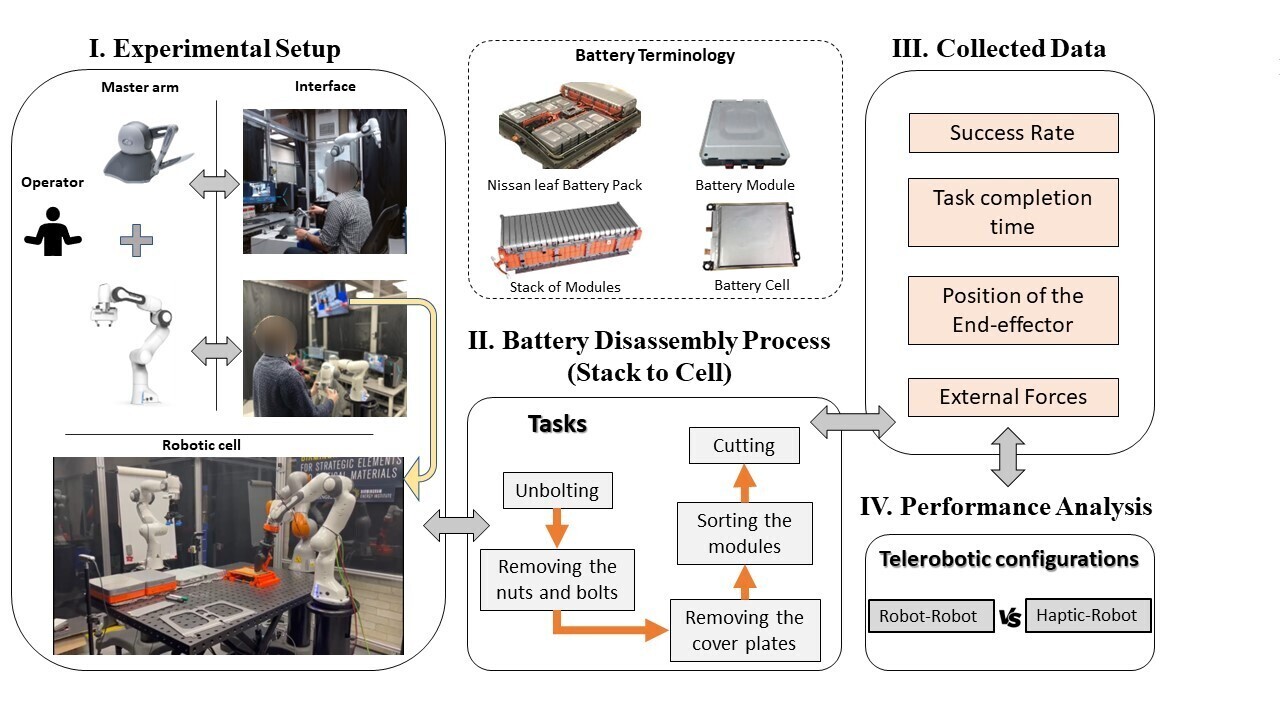}
    \caption{Overview diagram representing the experimental setup adopted in this work, alongside a brief breakdown for an EV battery disassembly process, from the battery pack to the battery cell. }\label{fig:Overview}
\end{figure}

This study focuses on the Nissan Leaf 2011 battery pack as a case study to demonstrate a range of disassembly tasks using telemanipulation. The Nissan Leaf 2011 pack comprises 192 cells enclosed within 48 modules, arranged with two forward vertical stacks of 12 modules and a single rear horizontal stack of 24 modules. An overview of the main disassembly stages for pack-to-cell disassembly is given in Table \ref{tab:ExpSetup-DisassemblyTable}; for brevity, this is provided for only the forward module stacks. The level of autonomy presented in Table \ref{tab:ExpSetup-DisassemblyTable} was inferred from understanding of the current manual disassembly process and use of design features, such as types of fasteners and accessibility of components. In general, the range of semi-/fully autonomous tasks will vary between battery designs and manufacturers, however, overall, there is a motivation to increase the level of autonomy to decrease disassembly costs \cite{LANDER2023120437}. We formulate a basic battery disassembly workstation design consisting of 2 Franka Emika Panda cobots and a Phantom Omni haptic device. The Franka Emika Panda is a 7 degree of freedom (DoF) robot with a 3kg payload. A combination of torque control and onboard torque sensing capabilities and low payload makes this robot suitable for accomplishing precise assembly and disassembly tasks in a shared human-robot workspace. The Phantom Omni comprises a 6 DoF handheld articulated arm with independent control of force feedback along 3 axes, with a maximum force capability of 3.3N. All devices were connected in a ROS network to a desktop with Intel(R) Core(TM) i7-8086K 8-core processor with 4 GHz base clock and 32 GB RAM. An overview of the experimental setup is shown in Figure \ref{fig:Overview}.

\begin{figure}
    \centering
    \begin{minipage}[b]{0.23\textwidth}
        \centering
        \includegraphics[angle=45,origin=c,width=\textwidth]{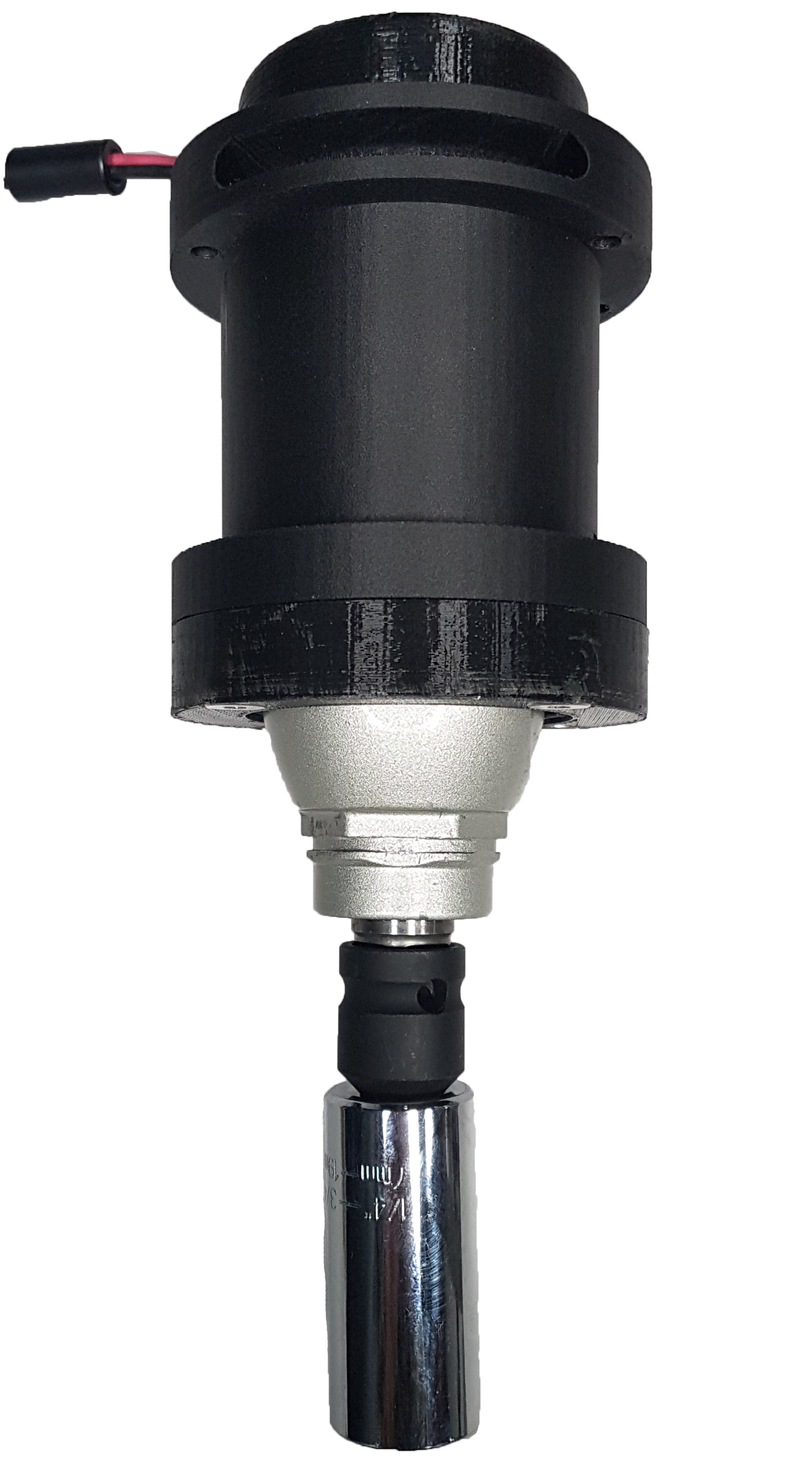}\\
        (a) Socket wrench tool
    \end{minipage}\hfill
    \begin{minipage}[b]{0.23\textwidth}
        \centering
        \includegraphics[angle=45,origin=c,width=\textwidth]{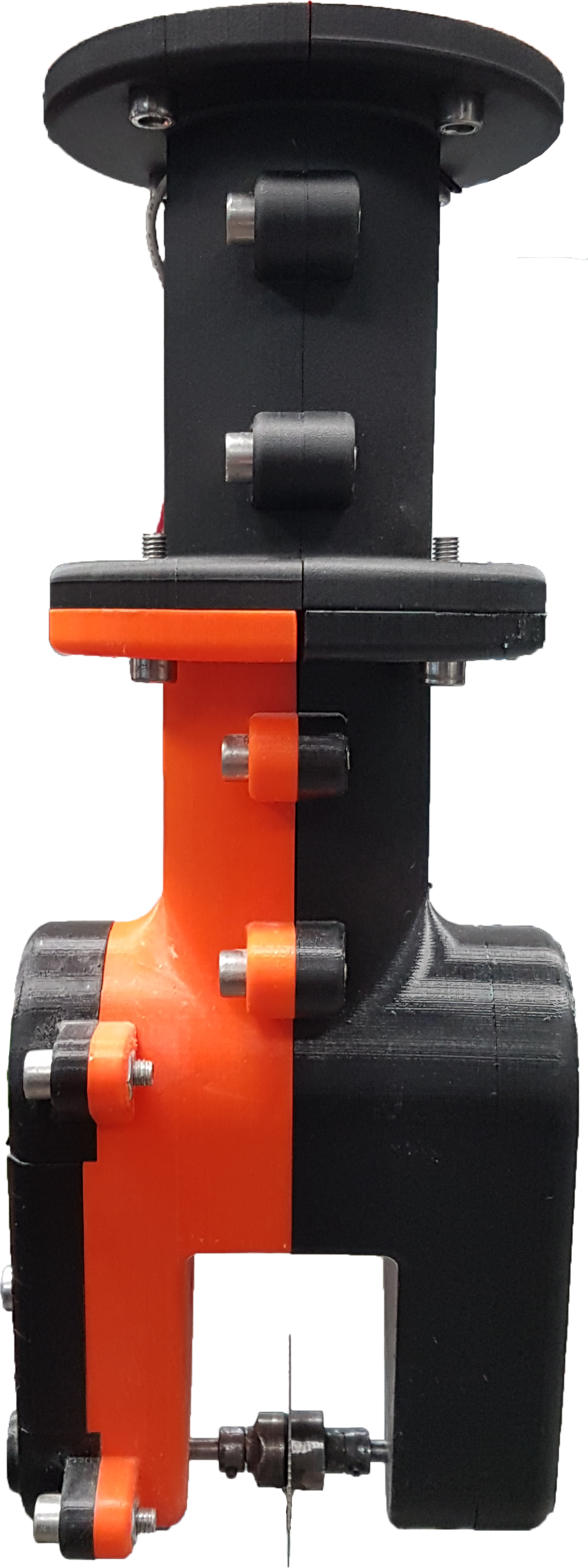}\\
        (b) Motorised cutter tool
    \end{minipage}\hfill
    \begin{minipage}[b]{0.23\textwidth}
        \centering
        \includegraphics[angle=45,origin=c,width=\textwidth]{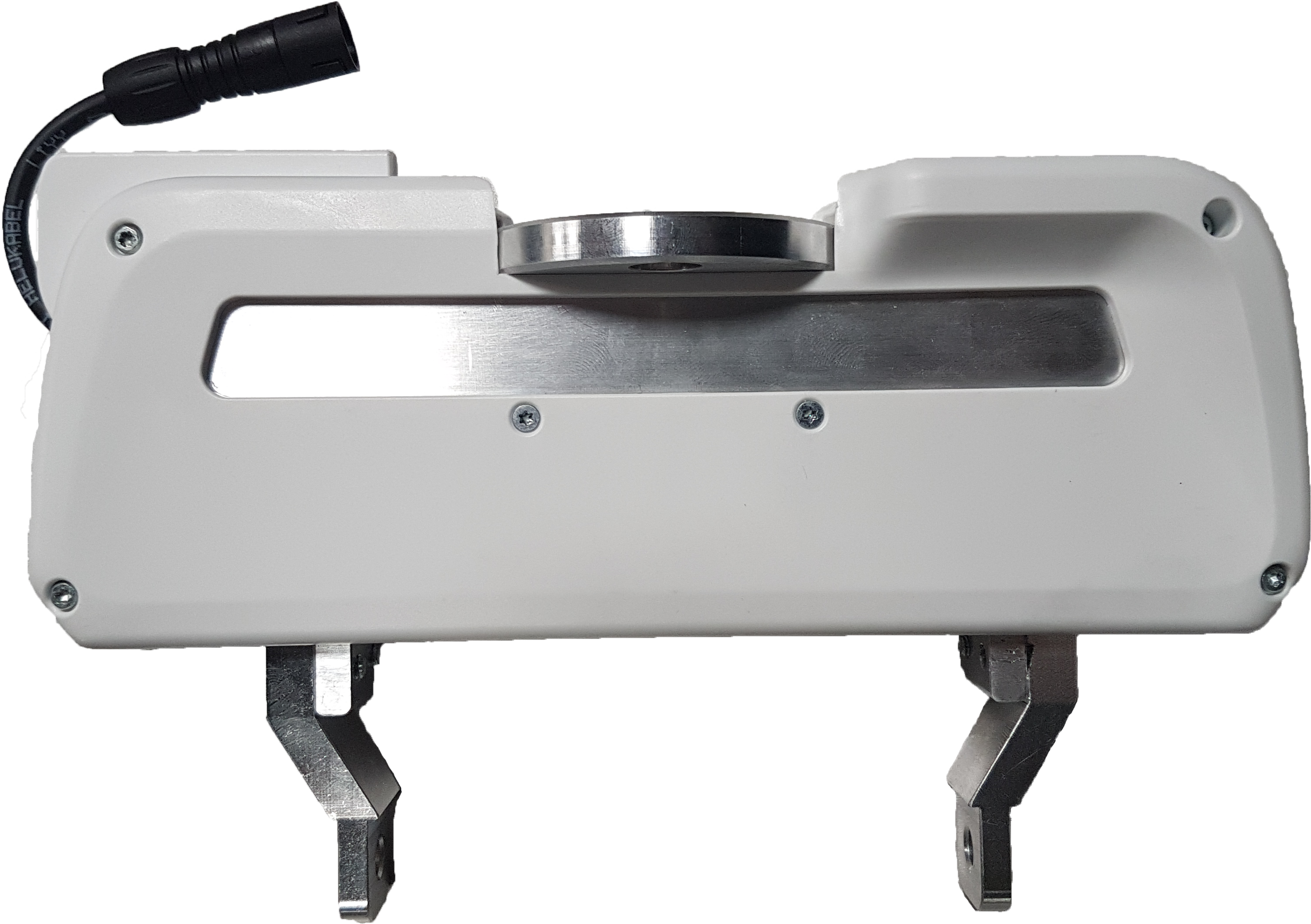}\\
        (c) Two-finger gripper
    \end{minipage}\hfill
    \begin{minipage}[b]{0.23\textwidth}
        \centering
        \includegraphics[angle=45,origin=c,width=\textwidth]{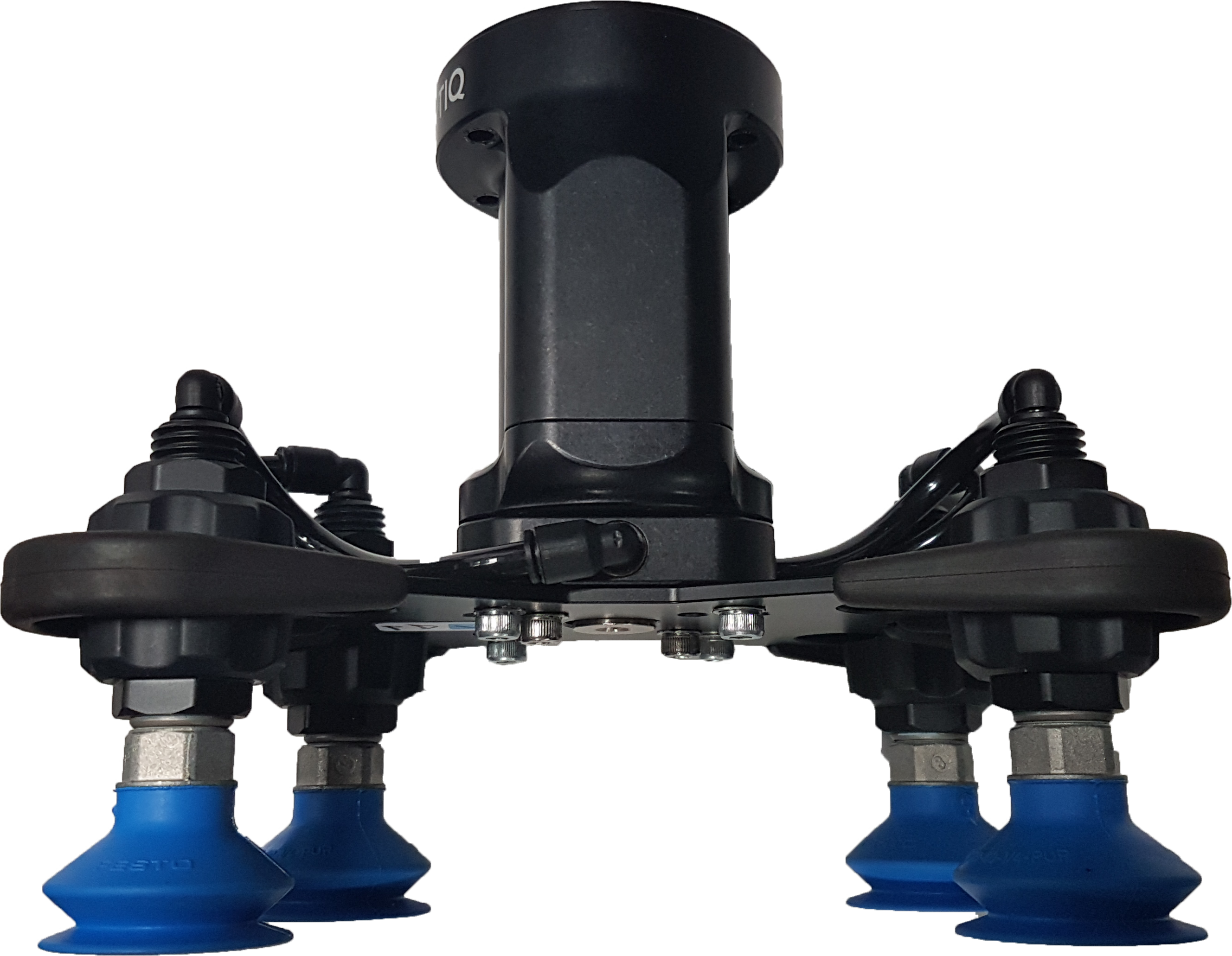}\\
        (d) Vacuum gripper
    \end{minipage}
    \caption{Tool attachments considered for the tasks of unbolting, cutting and sorting.}
    \label{fig:ExpSetup-Tools}
\end{figure}

Based on the disassembly sequence in Table \ref{tab:ExpSetup-DisassemblyTable}, we consider a selection of repetitive tasks identified in related studies \cite{HybridDisassemblyEVBatteries} where robots provide an advantage over manual disassembly. These consist of unbolting, such as of fasteners connecting a stack of modules; removal and sorting (pick and place) of disassembled waste components, and cutting to mechanically separate components where fasteners cannot be removed non-destructively. To deal with the range of tasks presented in the case of battery disassembly, a range of commercially available and custom designed tools were employed, shown in Figure \ref{fig:ExpSetup-Tools}. For this study, we design a custom socket wrench tool that can be employed with a range of fastener sizes. For cutting, a motorised slitting saw tool was designed specifically for the low-payload cobots presented to accomplish low-power cutting tasks. For pick and place, the Franka Hand two-finger gripper was employed for grasping thin, light wastes such as bolts and plates, while the Robotiq EPick suction gripper is applied for larger, heavier materials such as battery modules.

\begin{table}[]
    \small\centering
    \begin{tabular}{>{\raggedright}p{0.06\textwidth}>{\raggedright}m{0.16\textwidth}>{\raggedright}p{0.64\textwidth}>{\raggedright}p{0.05\textwidth}}
    \hline 
    Step \# & \multicolumn{2}{l}{Disassembly task} & Type\tabularnewline
    \hline 
    1 & \multirow{2}{0.16\textwidth}{Top case} & Remove service plug retainer & M\tabularnewline
    \cline{3-4} \cline{4-4} 
    2 &  & Remove upper case bolts \& lift top case & At\tabularnewline
    \hline 
    3 & \multirow{2}{0.16\textwidth}{Battery controller} & Remove mounting bolts & S-At\tabularnewline
    \cline{3-4} \cline{4-4} 
    4 &  & Disconnect harness connectors \& remove battery controller & M\tabularnewline
    \hline 
    5 & \multirow{11}{0.16\textwidth}{Junction box \& harnesses} & Disconnect interlock circuit harness \& heater harness connectors & M\tabularnewline
    \cline{3-4} \cline{4-4} 
    6 &  & Remove mounting nuts \& front stack connecting bus-bar & S-At\tabularnewline
    \cline{3-4} \cline{4-4} 
    7 &  & Remove battery member pipe & S-At\tabularnewline
    \cline{3-4} \cline{4-4} 
    8 &  & Remove junction box cover & M\tabularnewline
    \cline{3-4} \cline{4-4} 
    9 &  & Remove central bus bar bolts and remove central bus bar & S-At\tabularnewline
    \cline{3-4} \cline{4-4} 
    10 &  & Remove current sensor bus bar mounting bolt & S-At\tabularnewline
    \cline{3-4} \cline{4-4} 
    11 &  & Remove switch bracket mounting bolts & S-At\tabularnewline
    \cline{3-4} \cline{4-4} 
    12 &  & Invert switch bracket, disconnect harnesses \& remove switch bracket & M\tabularnewline
    \cline{3-4} \cline{4-4} 
    13 &  & Remove high voltage (HV) harness bolts \& remove HV harnesses & S-At\tabularnewline
    \cline{3-4} \cline{4-4} 
    14 &  & Disconnect voltage \& temperature sensor harnesses & M\tabularnewline
    \cline{3-4} \cline{4-4} 
    15 &  & Remove junction box mounting nuts \& junction box & M\tabularnewline
    \hline 
    16 & \multirow{3}{0.16\textwidth}{Heaters} & Disconnect harness connectors from heater and heater relay unit & M\tabularnewline
    \cline{3-4} \cline{4-4} 
    17 &  & Remove heater \& heater relay mounting nuts & S-At\tabularnewline
    \cline{3-4} \cline{4-4} 
    18 &  & Remove heater controller unit \& heaters & M\tabularnewline
    \hline 
    19 & \multirow{7}{0.16\textwidth}{\emph{Front module stack(s)}} & Remove stack mounting nuts & At\tabularnewline
    \cline{3-4} \cline{4-4} 
    20 &  & Extract module stack & At\tabularnewline
    \cline{3-4} \cline{4-4} 
    21 &  & Remove bus bar cover & M\tabularnewline
    \cline{3-4} \cline{4-4} 
    22 &  & Remove bus bar terminal mounting bolts \& mounting screws & S-At\tabularnewline
    \cline{3-4} \cline{4-4} 
    23 &  & \emph{Remove end plate bolts} & At\tabularnewline
    \cline{3-4} \cline{4-4} 
    24 &  & \emph{Remove end plate} & S-At\tabularnewline
    \cline{3-4} \cline{4-4} 
    25 &  & \emph{Electrical test \& sort modules} & S-At\tabularnewline
    \hline 
    26 & \multirow{3}{0.16\textwidth}{\emph{Module}} & \emph{Separate module cover} & S-At\tabularnewline
    \cline{3-4} \cline{4-4} 
    27 &  & Glue separation & At\tabularnewline
    \cline{3-4} \cline{4-4} 
    28 &  & Separate cell tabs from terminal assembly & S-At\tabularnewline
    \hline 
    \end{tabular}
    \caption{Sequence of disassembly operations for pack-to-cell disassembly of Nissan Leaf 2011. For brevity, only disassembly of the front module stack is considered. The type of task can be classified as fully manual (M), requiring specialised tools or dexterous hand-manipulation, semi-autonomous (S-At), where a robot can accomplish the task with the assistance of a human operator, and fully autonomous (At) tasks accomplishable without human intervention. A contiguous sequence of semi-autonomous processes can be observed for disassembly and sorting of the module stack; tasks considered in this study are \emph{emphasised}.}
    \label{tab:ExpSetup-DisassemblyTable}
\end{table}

\subsection{Master-Slave Telemanipulation using a Haptic Device as Master and a Cobot as Slave}

\begin{figure}
    \centering
    \includegraphics[width=0.95\textwidth]{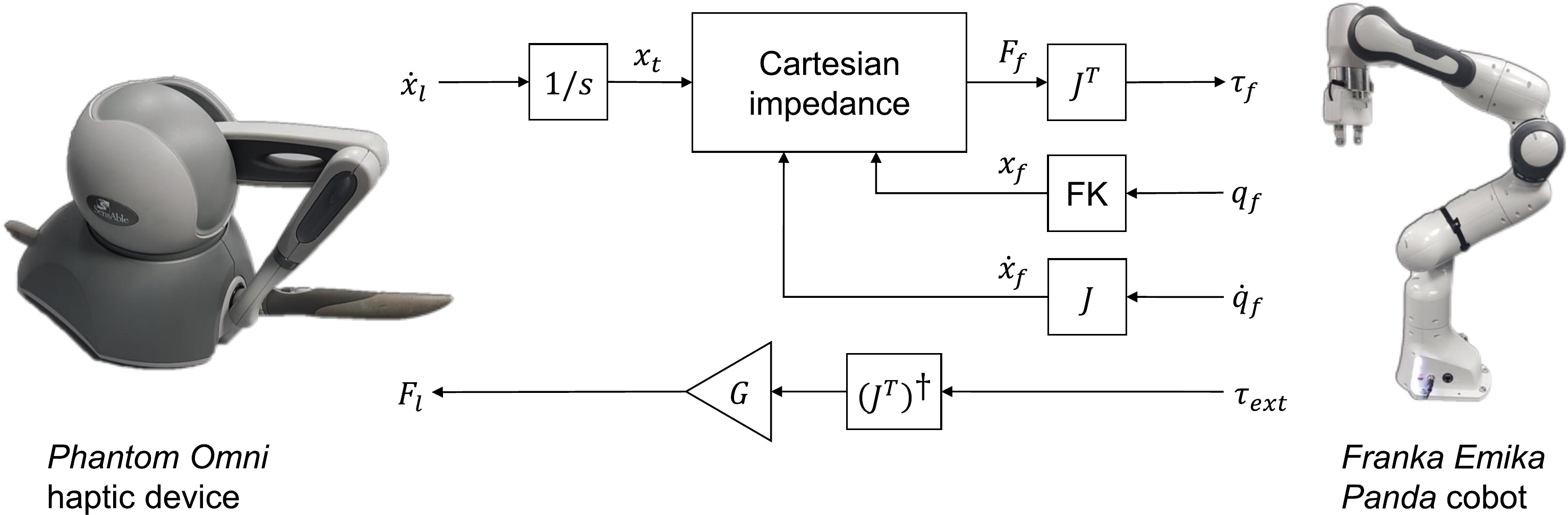}
    \caption{Outline of telemanipulation framework consisting of Phantom Omni haptic device and Franka Emika Panda cobot. FK refers to the forward kinematic mapping of follower joint configuration $\vec{q}_{f}$ to end-effector pose $\vec{x}_{f}$. Note the transformation matrices relating the base frames of the haptic and cobot are omitted for simplicity.}\label{fig:ExpSetup-HapticFranka}
\end{figure}

A common framework to accomplish telemanipulation tasks is to use a lower-cost, handheld platform such as a haptic device in combination with a robot in a master-slave configuration. In this case, we consider the case of a Phantom Omni haptic device and a Panda cobot (Figure \ref{fig:ExpSetup-HapticFranka}). Due to a mismatch in the DoFs of the Omni and Panda robot - more generally, any over-actuated robot - there is a limitation that not all degrees of freedom of the robot can be controlled independently, reducing the amount of control provided to the user. Typically, this requires a definition of a mapping function that maps the joint positions of the haptic device to that of the slave arm. Such a mapping can be achieved by mirroring the Cartesian pose or twist of the haptic end-effector to that of the slave robot. This provides a more intuitive framework for the user to position the robot to accomplish tasks. However, this typically renders the control vulnerable to singularities and joint limits, due to the solution of inverse kinematics to compute the required joint-space motion. Furthermore, the smaller scale of motion afforded by the Omni in both joint and task space implies a compromise between speed of coarse motion, where the motion of the haptic device is scaled to achieve larger motions on the robot; or precision of fine positional alignment, where the motion of the haptic device is mapped directly to the robot or even reduced. Moreover, due to the limited force capabilities of the haptic device, the force feedback cannot be mapped from the robot to the haptic 1:1. Therefore, the feedback can potentially feel unnatural or deliver insufficient cues for the operator, such as when exceeding safety limits imposed on force.

\subsection{Master-Slave Telemanipulation Using Two Identical Cobots}

Alternatively, in this work we consider a platform consisting of two identical Franka Emika Panda cobots operating in a master-slave configuration. With this approach, the user directly manipulates the master Franka arm, whose motions are directly mirrored to that of the slave arm. Owing to the identical configuration of both robots, both joint position and torques can be mapped between the robots 1:1. This 1:1 mapping results in natural and responsive feedback being delivered to the user. However, this also poses hazards as the user will be potentially exposed to the full forces involved in a specific task. The joint space control of both arms furthermore has advantages in the problem of singularities and joint limits, which are present in mapping the Cartesian end-effector pose of the master to the task space pose of the slave robot. However, the lack of task-space control has potential drawbacks when accomplishing tasks along specific task directions, such as removing bolts or cutting, more difficult and less intuitive for the user.

\begin{figure}
    \centering
    \includegraphics[width=0.95\textwidth]{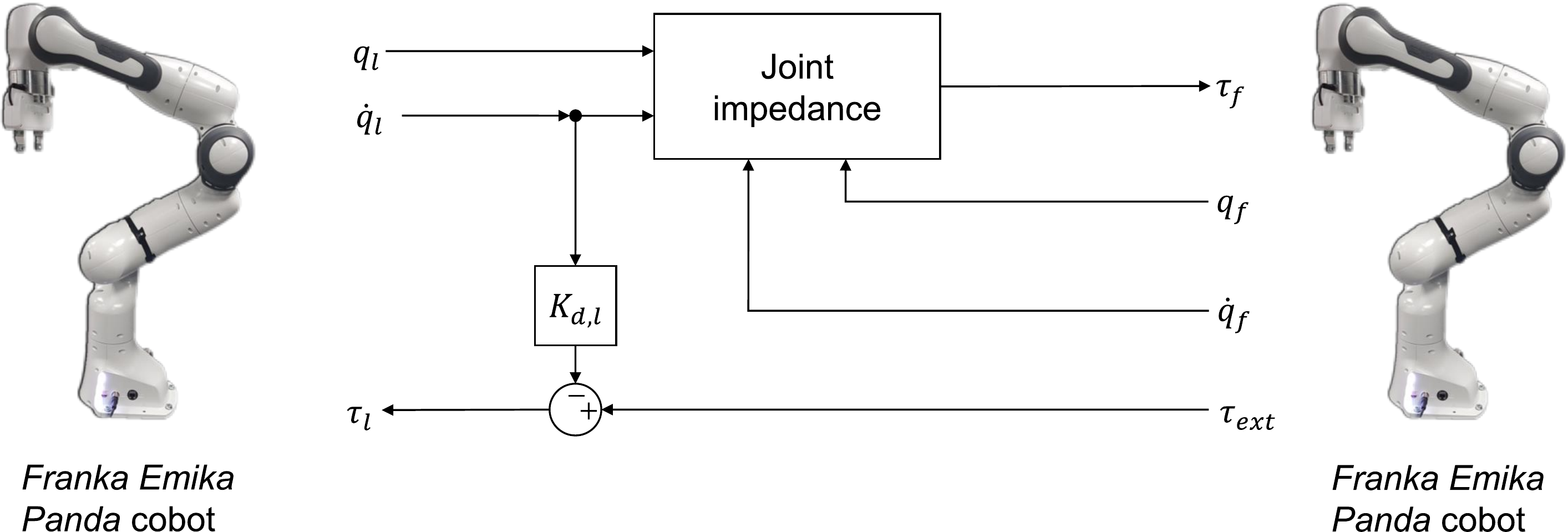}
    \caption{Outline of telemanipulation framework consisting of two identical Franka Emika Panda cobots.}\label{fig:ExpSetup-TwoFrankas}
\end{figure}

\section{Methodology} \label{sec:Methodology}

To achieve a virtual coupling between master and slave devices, we first consider the dynamic equation of a rigid $N$-link manipulator in joint space
\begin{equation}\label{eq:Method-Robot-Dynamics}
    \mat{M}(\vec{q})\vec{\ddot{q}} + \mat{C}(\vec{q},\vec{\dot{q}}) + \vec{g}(\vec{q}) = \vec{\tau}_{\mathrm{ext}} + \vec{\tau}
\end{equation}
where $\mat{M}\in\mathbb{R}^{N\times N}$, $\mat{C}(\vec{q},\vec{\dot{q}})\in\mathbb{R}^{N\times N}$, $\vec{g}(\vec{q})\in\mathbb{R}^{1\times N}$ are the joint-space inertia matrix, Coriolis and centrifugal matrix and gravitational torques respectively, and $\vec{\tau}_{\mathrm{ext}}\in\mathbb{R}^{1\times N}$, $\vec{\tau}\in\mathbb{R}^{1\times N}$ are the vectors of external and control torques acting on each link respectively. We denote the joint configurations $\vec{q}$ and command torques $\vec{\tau}$ by subscripts $l,f$ for master and slave respectively.

\begin{figure}[ht]
\centering
\includegraphics[width=0.75\textwidth]{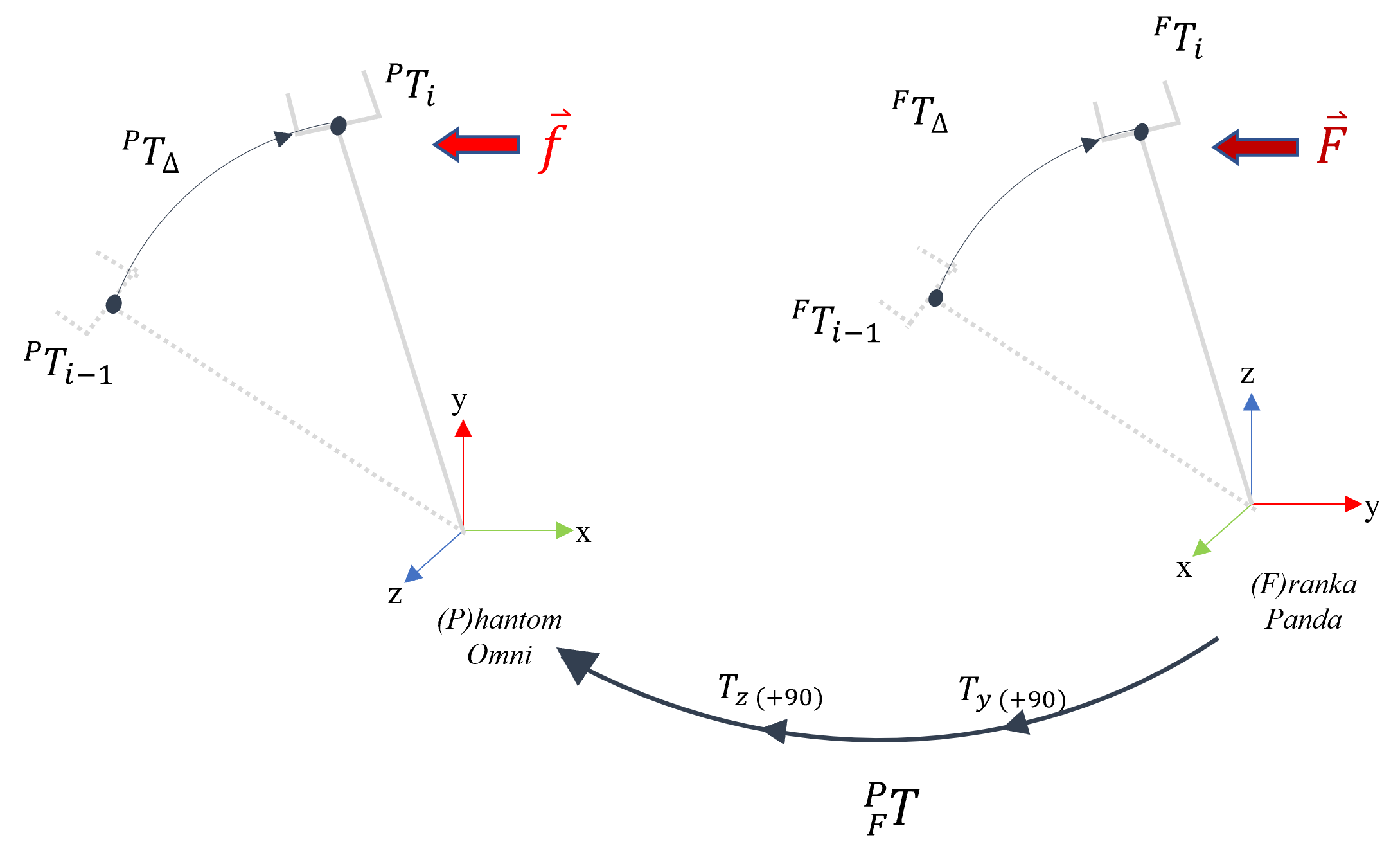}
\caption{Cartesian mapping for the master-slave telemanipulation setup between the \textit{Phantom Omni} haptic device and \textit{Franka Panda} robot arm.}\label{fig:mapping}
\end{figure}

For the haptic device, 1:1 mapping between each joint cannot be achieved in practice because of its 6 DoF kinematic chains it has compared to the 7 DoF Franka arm. Hence, it is required to either map the joint space onto a reduced subset of the full joint space of the robot, or operate in Cartesian space. In this work, Cartesian mapping is employed to velocity control the slave Franka arm with the Phantom Omni master arm in all Cartesian motion axes (XYZ translation and rotation). As the human operator moves the master arm in its workspace, delta Cartesian pose $\vec{\dot{x}_{l}}$, composed of position and orientation, is computed and then commanded to the slave arm. However, since reference coordinate systems of both arms are different (as shown in Figure \ref{fig:mapping}), delta Cartesian pose $\vec{\dot{x}_{l}}$ undergoes transformations to correctly map it with respect to Franka arm's base frame. The transformation mapping is calculated as
\begin{equation}\label{eq:mapping}
    {^F}\mat{T}_{\Delta} = {}^{P}_{F}\mat{T} \cdot {^P}\mat{T}_{\Delta} = \left[\begin{array}{c c} 
                                        	{^F}\mat{R}_{\Delta} & {^F}\vec{t}_{\Delta}\\ 
                                        	\hline 
                                        	0 & 1 
                                            \end{array}\right]
\end{equation}
where ${}^{P}_{F}\mat{T}$ is the homogeneous transformation matrix from Franka arm base frame to Phantom Omni base frame, ${^F}\mat{T}_{\Delta}$, and ${^P}\mat{T}_{\Delta}$ represent the delta transformation matrices for Franka and Phantom respectively. Position and orientation components of ${^P}\mat{T}_{\Delta}$ are assigned to the desired pose $\vec{x}_{t}$ to compute the pose error $\vec{e}_{x}=\vec{x}_{f}-\vec{x}_{t}$ for the slave arm. Thus, the slave arm control law is:
\begin{equation}
    \vec{\tau}_{f} = \mat{J}^{\tpose}\left(-\mat{K}_{p}\vec{e}_{x} - \mat{K}_{d}\mat{J}\vec{\dot{q}_{f}} \right) + \mat{C}(\vec{q_{f}}) + \vec{g}(\vec{q_{f}})
\end{equation}
where $\mat{J}\in\mathbb{R}^{7\times 6}$ is the slave manipulator Jacobian mapping joint to end-effector velocities and $\mat{K}_{p}$, $\mat{K}_{d}$ are controller stiffness and damping matrices respectively. This results in the desired closed-loop dynamic behaviour from \eqref{eq:Method-Robot-Dynamics} (where $\mat{\Lambda}\in\mathbb{R}^{6\times6}$ is the operational space inertia matrix):
\begin{equation}
    \mat{\Lambda}\vec{\ddot{x}_{f}} + \mat{K}_{d}\vec{\dot{x}_{f}} + \mat{K}_{p}\vec{e}_{x} = \vec{F}_{\mathrm{ext}}
\end{equation}

\textbf{Force Feedback} is an essential part of a bilateral teleoperation system in which a force feedback is maintained at control frequency. This helps the user to have a tactile perception of the slave robot's environment. In a similar way to \eqref{eq:mapping}, external force vector $\vec{F}_{\mathrm{ext}}$ experienced at the Franka end-effector is transformed with respect to the Phantom so the user is able to perceive it as force $\vec{F}_{l}$ as follows:
\begin{equation}\label{eq:fmapping}
    \vec{F}_{l} = G\cdot{^P _F}\mat{T}^{-1} \cdot \vec{F}_{\mathrm{ext}}
\end{equation}
Due to the mismatch in force capabilities between master and slave devices, the feedback is scaled by a factor $G=0.1$. This factor was determined experimentally by comparing the maximimum expected force across all tasks from preliminary data and scaling to the maximum force capabilities of the haptic device (3.3N). To distort the force feedback to a minimal extent, and to maintain consistency with the constant 1:1 feedback of the identical cobot setup, this factor was held constant across all trials.

For the Franka arm, a joint impedance control scheme is used that directly maps the joint configuration of the master arm to the slave arm (Figure \ref{fig:ExpSetup-TwoFrankas}). The control law is defined as follows:
\begin{equation}
    \vec{\tau}_{f} = -\mat{K}_{p}\vec{e}_{q} - \mat{K}_{d}\vec{\dot{e}}_{q} + \mat{C}(\vec{q}_{f}) + \vec{g}(\vec{q}_{f}) 
\end{equation}
where $\vec{e}_{q}=q_{f}-q_{l}$ is the joint space error. This results in the closed-loop dynamics in joint space for the slave arm
\begin{equation}
    \mat{M}\vec{\ddot{q}_{f}} + \mat{K}_{d}\vec{\dot{e}}_{q} + \mat{K}_{p}\vec{e}_{q} = \vec{\tau}_{\mathrm{ext}}
\end{equation}
To provide force feedback to the user, the master control torques are calculated as:
\begin{equation}
    \vec{\tau}_{l} = \vec{\tau}_{\mathrm{ext}} - \mat{K}_{d,l}\vec{\dot{q}}_{l}
\end{equation}
where the force feedback is computed directly as the estimated external torque applied to the slave arm. Due to the combination of higher forces applied to the master with the low stiffness of the human operator, an additional damping term $\mat{K}_{d,l}$ is added to the response of the master to reduce oscillations resulting from the displacement of the master arm caused by the bilateral master-slave coupling.

\section{Case Studies} \label{sec:Case Studies}

\begin{figure}[h]
    \centering
    \includegraphics[width=1\textwidth]{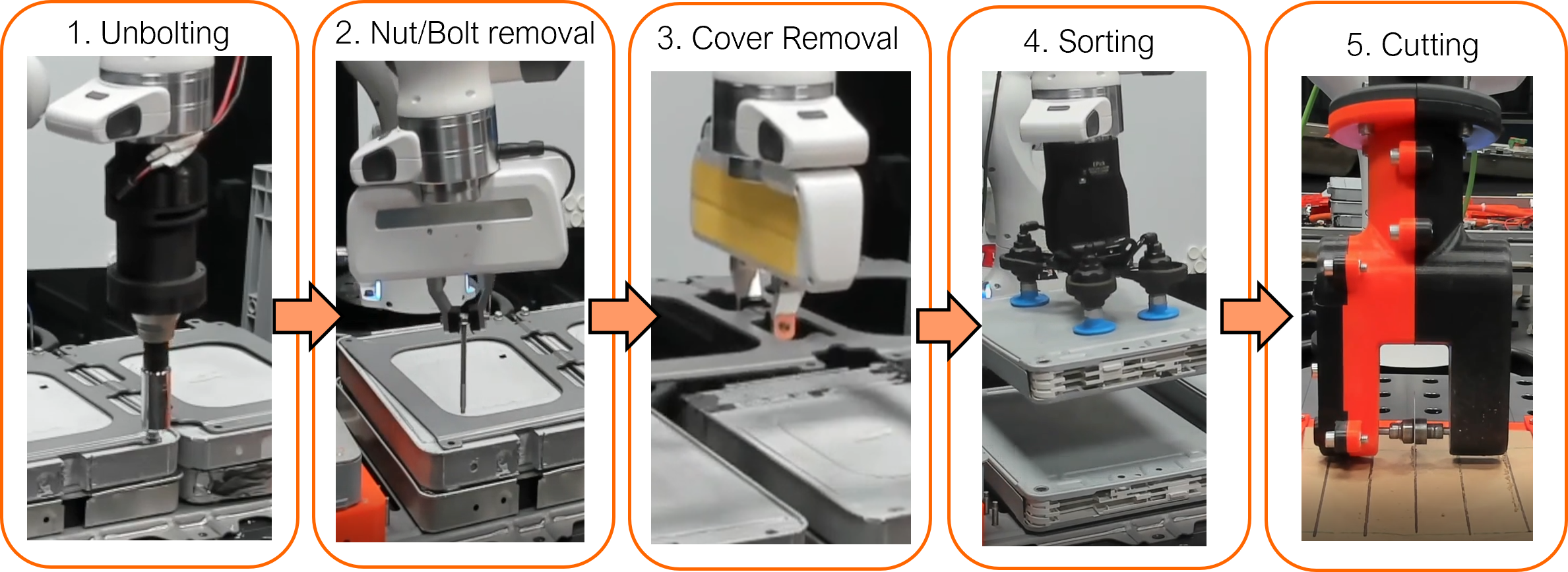}
    \caption{Part of an EV battery disassembly process showing the case studies experimented on through this work in order: 1) Unbolting fasteners. 2) Removing fasteners 3) Removing module cover plate. 4) Sorting modules. 5) Surface contact cutting.}\label{fig:CaseStudies}
\end{figure}

A contiguous sequence of tasks relating to the disassembly and extraction of a stack of modules from the Nissan Leaf 2011 battery pack were identified as feasible targets for robotic disassembly, presented in Figure \ref{fig:CaseStudies}. The considered case study encapsulates a range of contact-rich tasks required to disassemble and recover the battery modules from an individual stack. The modules contain the active material within the battery and hence are considered a safety-critical aspect of the disassembly process due to the associated handling risks, such as risk of shorting the terminals and risk of mechanical shock from mishandling. Each task was attempted five times (trials). For each task, the operator is provided with two fixed camera views of the module stack, and otherwise does not have a direct view of the scene. These views are held constant for all tasks with the exception of cutting, where the operator is instead provided two camera views of a material holder containing the cutting workpiece. Prior to commencing each task, both master and slave robots are initialised to a home position in joint space held constant for all tasks. As the focus of this work is not on human factor analysis and subjective measures of performance based on end-user evaluations, but on objective measures of task performance, the operator is assumed to have familiarity and prior experience with both teleoperation platforms. In addition, each operator was given 10 minutes of training with each task with both telemanipulation systems before first attempting each task. Four expert operators participated in the study. For each trial, a single operator was selected randomly to perform the task. A more detailed description of each task is provided as follows.

\subsection{Unbolting}
This task considers the unbolting of a set of 4 fasteners constraining an individual stack of modules. The operator has access to a motorised universal socket wrench tool mounted at the robot's wrist. The operator is responsible for aligning the socket wrench with the bolt and operating the tool to unscrew the fastener. The task is considered successful if the bolt is removable by hand without any further unscrewing action. If the success condition is not met after the first attempt, or the configured force thresholds (40N) of the robot are exceeded, the task is considered a failure.

\subsection{Removing the Fasteners}
This case study follows from the previous unbolting stage by removing each fastener from the stack before removal of the cover. For this task, the Franka hand was used with a configured grasping force of 50N. The operator was required to remove a set of 8 fasteners by maneuvering the hand towards each of the fasteners, grasping and removing the bolt from the stack and depositing into a container. Due to the limitation of the camera views of the object, the operator must use a mixture of tactile and visual exploration to successfully remove the fasteners. For each bolt, the failure condition is met if the bolt is not grasped after the first grasp attempt, or if the grasp is lost outside of the target container.

\subsection{Removing the Module Cover Plate}
After the removal of the fasteners, the operator must then remove the module cover plate to access the underlying module stack. Due to the weight and geometry of the cover plate, it is essential for the operator to find and manoeuvre towards a good grasp point to ensure safe transportation of the cover during removal. The configured grasping force was increased to 60N for all experiments. The failure condition is met if the first grasp is unsuccessful or the grasp is lost during the transportation of the cover.

\subsection{Sorting Modules}
This case study considers unstacking and sorting of the EV battery modules using a vacuum suction gripper. The operator must remove a pair of modules from a stack and deposit them into a container. This task requires visual positioning of the gripper onto a suitable surface for grasping, while furthermore maintaining contact to engage the suction cups with the material without exceeding the force limits of the robot. Similarly to the cover removal case study, the failure condition is met if the first grasp is unsuccessful, or if the grasp is lost during transportation.

\subsection{Contact Cutting}
Additional semi-destructive disassembly tasks may be carried out to further disassemble the individual modules to access the Li-ion cells or to separate and remove fasteners or connectors that are not amenable to non-destructive disassembly. We consider the case study of contact cutting of a planar material along a predefined visually marked desired path using a slitting saw tool. Due to the limited availability of battery materials and the power limitations of the spindle, we consider cutting a cardboard sheet as a benchmark for the more general process of cutting different planar materials. The operator must use tactile feedback to control the cutting force while also using visual feedback to achieve precise positioning along the path. The desired cutting path consists of a window of $\pm$2.5mm measured from the centroid of the marked path. If the cut deviates from the path greater than this threshold, or if the cut is incomplete along parts of the desired path, the failure condition is met.

\section{Results and Discussion} \label{sec:ResultsDiscussion}

For each case study, the overall \textbf{success rate} and average task \textbf{completion time} over all trials are summarised in Table \ref{tab:Results-Task-Performance}. For all tasks, the success rate is observed to exceed or equal 50\% with both platforms, with the lowest success rate of 50\% observed for the task of bolt removal with an identical cobot. For the majority of the tasks, a success rate of over 75\% is achieved, demonstrating the feasibility of the module stack disassembly process with both cobot and haptic device tele-robotics platforms. Between each platform, the success rate is broadly comparable for unbolting and bolt removal, where the operator success rate is 10-15\% higher than the identical cobot case. However, cover removal is noticeably the most simple task as it is merely almost a planar trajectory; given that the operator managed to reach a firm grasp point on the cover plate, which was achieved 100\% in all operators' trials with both platforms. On the other hand, the largest differences are observed with unstacking and cutting, where in the former case the identical cobot is outperformed by the haptic device by 30\%, whereas in the latter case, the inverse is true, with a 20\% improvement in success rate. Unstacking findings can be justified due to the design mechanism of the vacuum suction gripper, where the gripper has to be in a direct contact and, more importantly, perpendicular orientation with respect to the module surface to achieve a successful grip. Hence, manipulating the suction gripper's position/orientation to achieve this in Cartesian space with the haptic platform showed higher success rate than in joint space with the identical cobot platform, where gripping failure cases occurred due to orientation misalignment. Furthermore, the findings of cutting trials emphasize the significance of force feedback. Identical cobot platform showed higher success rate because the force feedback is maintained at a 1:1 scale, whilst in haptic platform, it is scaled down due to limited force capabilities as \eqref{eq:fmapping}. Therefore, the user did not experience the full scaled forces exerted on the end-effector while cutting.

\begin{table}
\begin{tabular}{lllll}
\hline 
 & \multicolumn{2}{c}{Haptic} & \multicolumn{2}{c}{Franka}\tabularnewline
\hline 
 & Success rate {[}\%{]} & Avg. task time {[}s{]} & Success rate {[}\%{]} & Avg. task time {[}s{]}\tabularnewline
\hline 
Unbolting (4$\times$)        & 95  & 188$\pm$23 & 85  & 124$\pm$13 \tabularnewline
Nut/bolt removal (8$\times$) & 63  & 713$\pm$89 & 50  & 410$\pm$63 \tabularnewline
Cover removal                & 100 & 101$\pm$15 & 100 & 70$\pm$6  \tabularnewline
Sorting modules (2$\times$)  & 90  & 179$\pm$19 & 60  & 77$\pm$5  \tabularnewline
Cutting                      & 60  & 122$\pm$26 & 80  & 95$\pm$18  \tabularnewline
\hline
\end{tabular}
\caption{Comparison of performance metrics for disassembly tasks for telemanipulation with Phantom Omni haptic device and identical cobot (Franka) platforms.}
\label{tab:Results-Task-Performance}
\end{table}

\begin{figure}
    \centering
    \begin{minipage}[b]{0.21\textwidth}
        \centering
        \includegraphics[width=\textwidth]{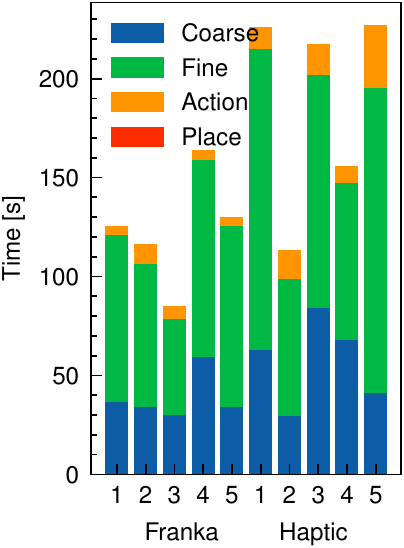}
        (a) Unbolting
    \end{minipage}\hfill
    \begin{minipage}[b]{0.19\textwidth}
        \centering
        \includegraphics[width=\textwidth]{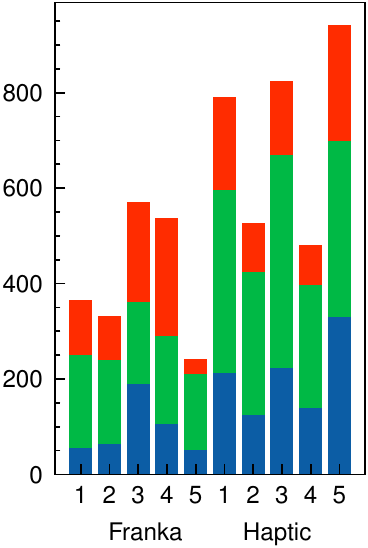}
        (b) Bolt removal
    \end{minipage}\hfill
    \begin{minipage}[b]{0.19\textwidth}
        \centering
        \includegraphics[width=\textwidth]{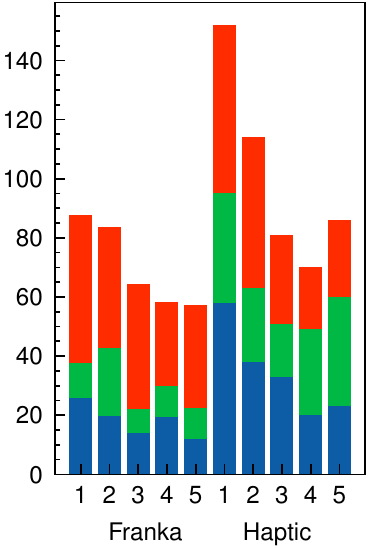}
        (c) Cover removal
    \end{minipage}\hfill
    \begin{minipage}[b]{0.19\textwidth}
        \centering
        \includegraphics[width=\textwidth]{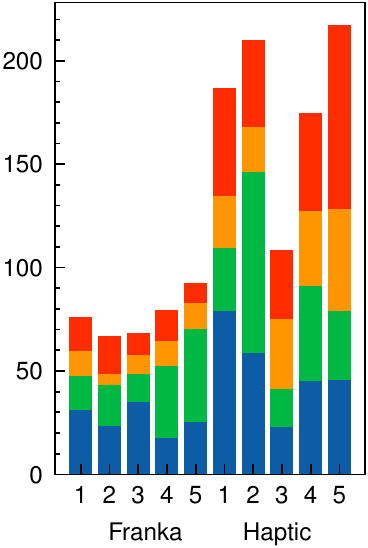}
        (d) Sorting modules
    \end{minipage}\hfill
    \begin{minipage}[b]{0.19\textwidth}
        \centering
        \includegraphics[width=\textwidth]{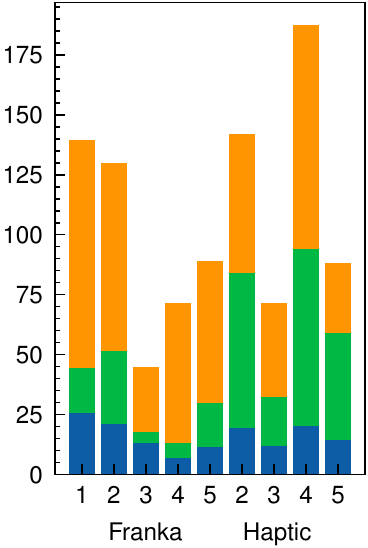}
        (e) Cutting
    \end{minipage}
    \caption{Breakdown by task trial of task completion times between Haptic and Franka telemanipulation masters. Data are omitted when the failure condition is met before completion of the task, such as by violation of configured force limits. Tasks are broadly split into 4 categories: an initial ``Coarse'' phase of approximate visual tool alignment, a ``Fine'' phase of precise visual and tactile alignment, an ``Action'' phase consisting of the task-relevant action, e.g. unbolting, cutting, and ``Place'' phase where the user deposits grasped objects.}
    \label{fig:Results-Times}
\end{figure}

In Figure \ref{fig:Results-Times}, a detailed breakdown of task completion times by trial is presented, indicating the time spent in each stage of the trials. Depending on the task, these stages subdivide each task consecutively, from start to completion, into categories: a ``Coarse'' stage where the operator telemanipulates the robot through rough visual alignment of the tool, followed by a ``Fine'' stage utilising tactile and visual feedback for precise positioning of the tool towards the interaction point. In the ``Action'' stage, the operator interacts with the component to perform a task after successful alignment of the tool. These actions correspond to unbolting, gripping and engagement of the cutter for nut/bolt removal, sorting and cutting respectively. Finally, the ``Place'' stage is encountered for grasping tasks after a successful grasp of the target object, during which the operator moves the grasped object and releases at a target position. The coarse and fine stages were annotated according to the proximity of the tool to the task-dependent target; a distance of the tool within 5cm of the target - for example, bolts and cover grasp points - indicates the fine alignment stage. The place stage was specified for coarse motions performed after grasping, which reverts to the coarse phase of motion after release of the grasped object. The action phase was annotated based on task-specific position and force thresholds. For example, for the sorting task, the task-relevant action is engagement and evacuation of the suction cups to grasp the module. From the 4 stages breakdown presented in Figure \ref{fig:Results-Times}, one can have a clear interpretation of what stages the operators spent most of the task time at, indicating their exerted efforts in completing each of the stages with respect to the two comparable platforms.

From an overview of Figure \ref{fig:Results-Times}, the identical cobot platform achieved shorter completion times on average across all tasks, compared to the haptic platform. Additionally, the highest difference in average time between the two platforms is observed for the bolt removal and sorting tasks, on average requiring $\sim$1.7--2.3$\times$ longer to complete with the haptic device respectively. Absolute task completion times were more consistent between trials with the identical cobot platform for all tasks. To evaluate the difference in task completion times in a timescale-invariant manner, the standardised mean difference (SMD) effect size metric was used. Given the mean and standard deviation $(\bar{t}_{h}, \sigma_{{t}_{h}})$, $(\bar{t}_{f}, \sigma_{{t}_{f}})$ for each platform respectively, the SMD can be defined as:
\begin{equation}
    \mathrm{SMD} = \frac{\bar{t}_{h}-\bar{t}_{f}}{\sqrt{\sigma_{t_{h}}^{2} + \sigma_{t_{f}}^{2}}}
\end{equation}
and provides an evaluation of the effect of each platform on completion time relative to the amount trial-to-trial variation. Values of $\left|\mathrm{SMD}\right|\geq0.8$ implies a significant effect, and the converse indicates a small or marginal effect.

For the considered module stack disassembly case study, the highest proportion of time was spent unbolting and removing the retaining fasteners. This is observed from Figure \ref{fig:Results-Times}a and \ref{fig:Results-Times}b, where the 'fine' stage occupies 60\%, 63\%, and 47\%, 50\% of the average task completion time, respectively, in both platforms; therefore, the fine alignment stage was the primary contributor to the reduction in overall task completion speed. This indicates that regardless of the platform used, these two tasks, among the other disassembly tasks, were the most effort demanding; requiring a combination of precise visual and tactile alignment of the tool to successfully unbolt/remove the fasteners. However, the SMD for unbolting and bolt removal respectively were 1.09 and 1.24, which indicates large improvements afforded by the identical cobot setup.

\begin{figure}
    \centering
    \begin{minipage}[b]{0.49\textwidth}
        \centering
        \includegraphics[width=\textwidth]{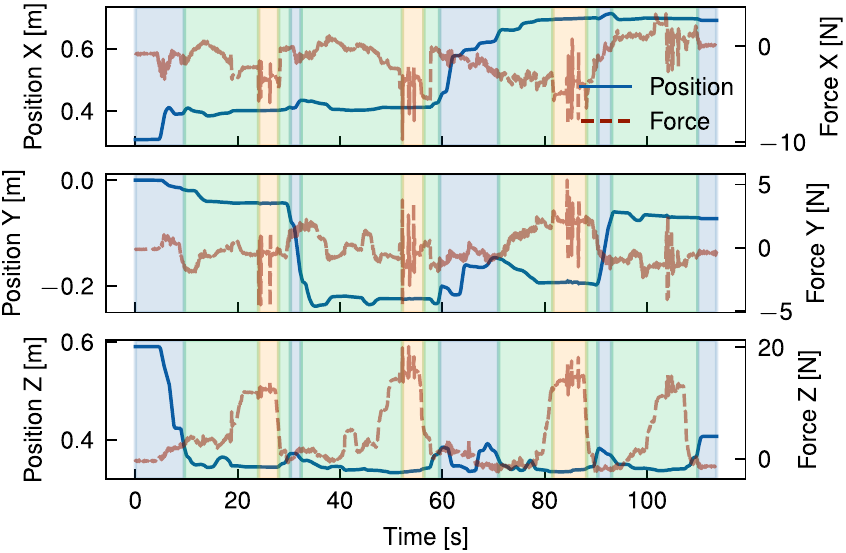}
        (a) Haptic master
    \end{minipage}\hfill
    \begin{minipage}[b]{0.49\textwidth}
        \centering
        \includegraphics[width=\textwidth]{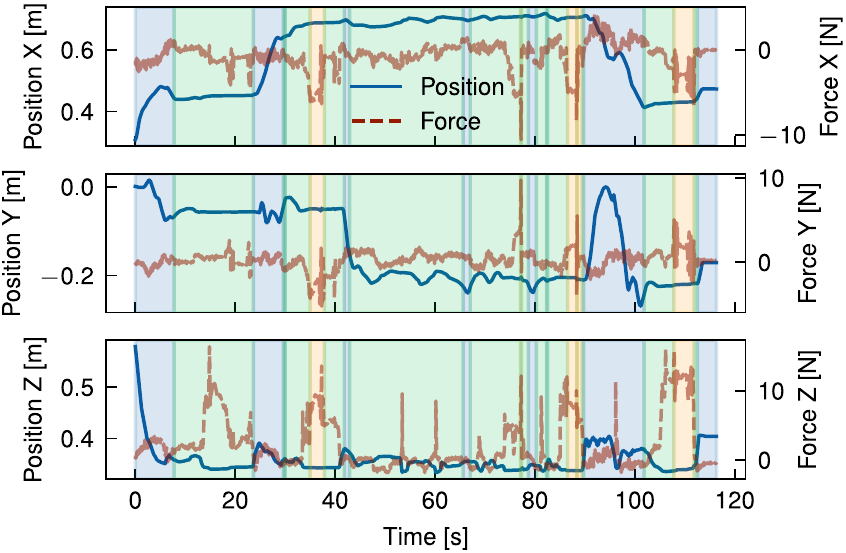}
        (b) Franka master
    \end{minipage}
    \caption{Overlay of position and estimated external forces at the end-effector for the \textbf{unbolting task} (trial 2) using a motorised universal socket wrench tool. Increases in force ($\sim$10-15N) in the surface normal direction (Z) indicate establishment of contact with the bolt and tactile exploration to identify correct alignment of the tool. Execution of unbolting is marked by rapid force fluctuations around these peaks.}
    \label{fig:Results-Unbolting-PosForce}
\end{figure}

\begin{figure}
    \centering
    \begin{minipage}[b]{0.49\textwidth}
        \centering
        \includegraphics[width=\textwidth]{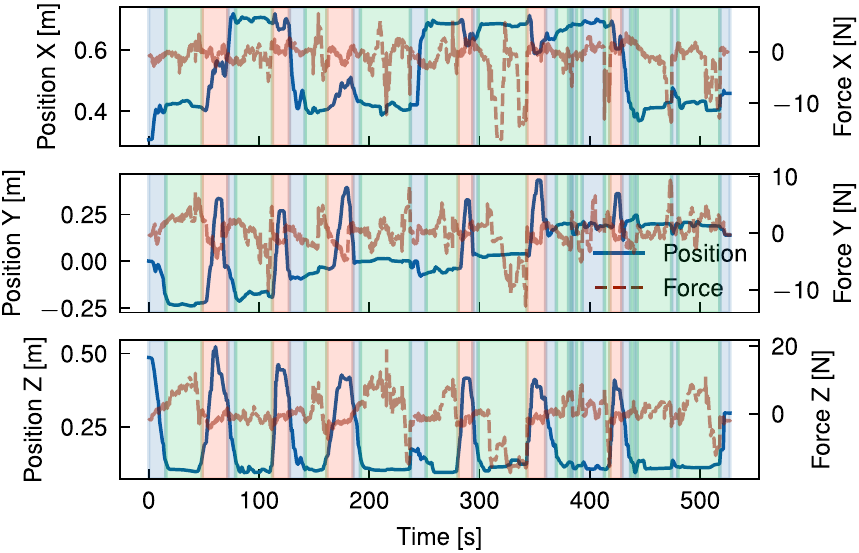}
        (a) Haptic master
    \end{minipage}\hspace{0.01\textwidth}
    \begin{minipage}[b]{0.49\textwidth}
        \centering
        \includegraphics[width=\textwidth]{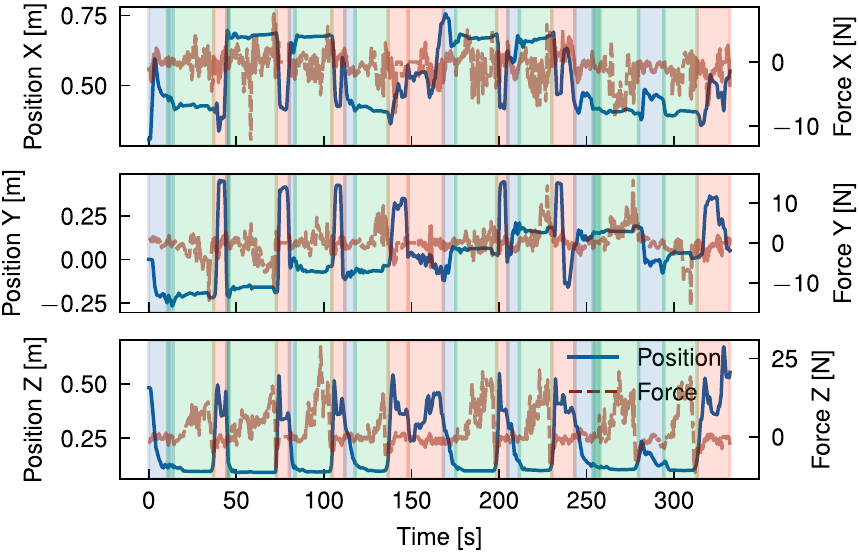}
        (b) Franka master
    \end{minipage}
    \caption{Overlay of position and estimated external forces at the end-effector for the \textbf{bolt removal} task (trial 2) using a 2-finger gripper. The position alignment can be categorised broadly into rapid visual alignment (indicated by a rapid decrease in Z position), followed by a slower phase of precise visual \& tactile alignment (plateau in Z position). Variations in force along the X-Y plane are indicative of tactile exploration to find suitable grasp points for the bolt.}
    \label{fig:Results-Rbolts-PosForce}
\end{figure}

\begin{figure}
    \centering
    \begin{minipage}[b]{0.49\textwidth}
        \centering
        \includegraphics[width=\textwidth]{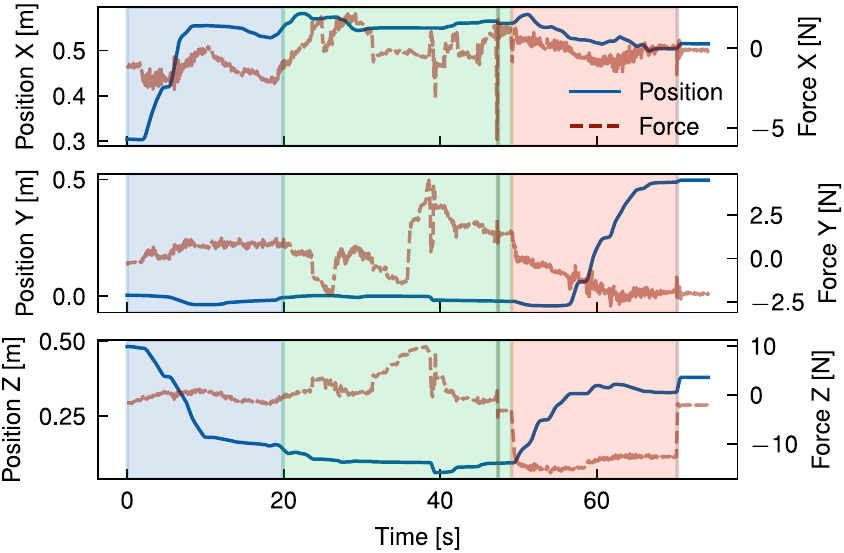}
        (a) Haptic master
    \end{minipage}\hfill
    \begin{minipage}[b]{0.49\textwidth}
        \centering
        \includegraphics[width=\textwidth]{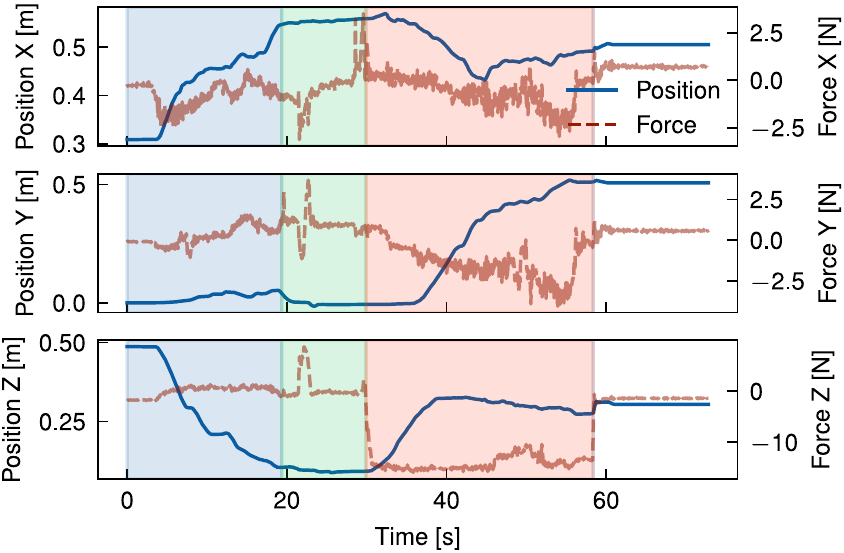}
        (b) Franka master
    \end{minipage}
    \caption{Overlay of position and estimated external forces at the end-effector for the \textbf{cover removal} task (trial 4) using a 2-finger gripper. An initial coarse alignment stage is marked by a rapid decrease in Z position, followed by a fine alignment guided by tactile feedback. Grasping and releasing of the cover is observed as application and release of a $\sim$14N force in the -Z direction.}
    \label{fig:Results-Rcover-PosForce}
\end{figure}

\begin{figure}
    \centering
    \begin{minipage}[b]{0.49\textwidth}
        \centering
        \includegraphics[width=\textwidth]{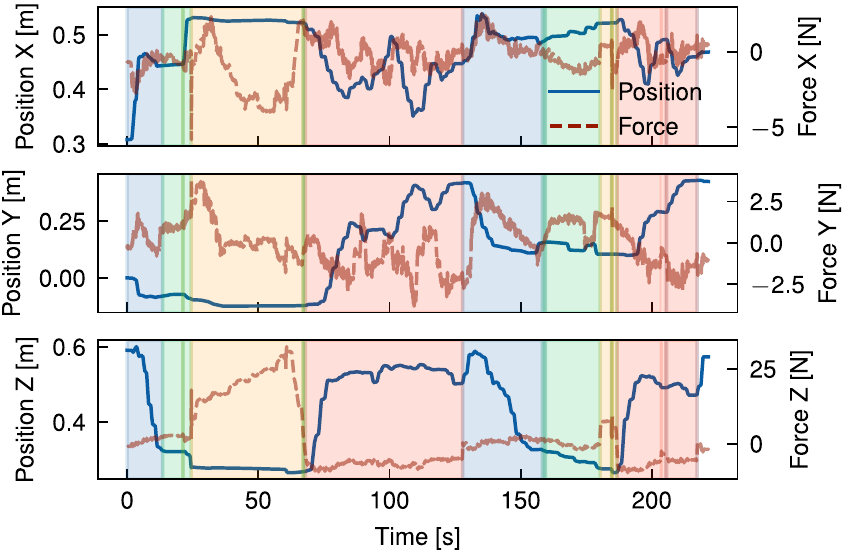}
        (a) Haptic master
        \label{fig:Results-Suction-PosForce-Haptic}
    \end{minipage}\hfill
    \begin{minipage}[b]{0.49\textwidth}
        \centering
        \includegraphics[width=\textwidth]{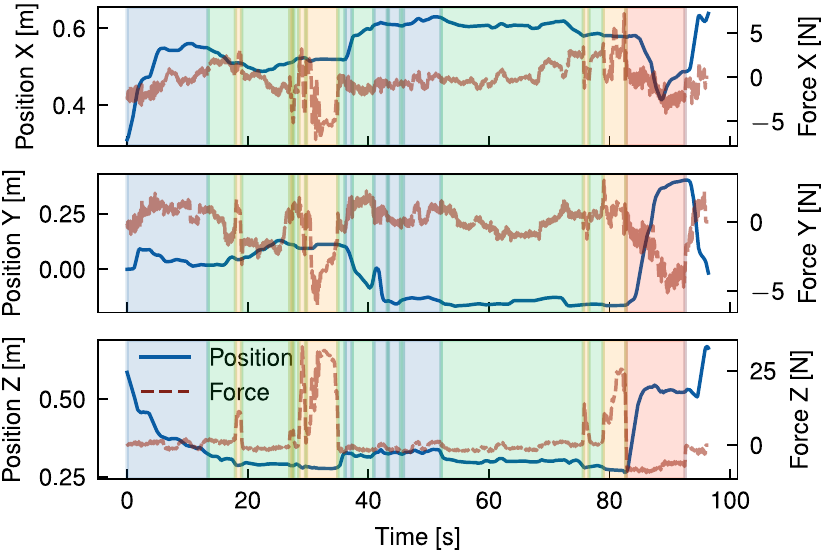}
        (b) Franka master
        \label{fig:Results-Suction-PosForce-Franka}
    \end{minipage}
    \caption{Overlay of position and estimated external forces at the end-effector for the \textbf{sorting modules} task (trial 5) using a vacuum suction gripper. An initial coarse alignment stage is marked by a rapid decrease in Z position, followed by a fine alignment guided by tactile feedback. Grasping and releasing of the module is observed as application and release of a $\sim$8N force in the -Z direction.}
    \label{fig:Results-Suction-PosForce}
\end{figure}

Figures \ref{fig:Results-Unbolting-PosForce}-\ref{fig:Results-Suction-PosForce} present the position graphs of the Franka slave's end-effector in Cartesian space, as well as the external forces experienced during a sampled trial from each task in the case study with both platforms. The figures also display the 4 categorised stages ('Coarse', 'Fine', 'Action', 'Place') color-shaded in each graph's background corresponding to each stage, as introduced in Figure \ref{fig:Results-Times}, throughout the time span of the presented trial. In Figure \ref{fig:Results-Unbolting-PosForce}, spikes in the force z plot enclosed in the yellow region ('action' stage) correspond to the action of \textit{unbolting} of the fasteners using the wrench tool. Similarly, Figure \ref{fig:Results-Rbolts-PosForce} represents the \textit{bolt removal} task in which fasteners are grasped, lifted up from their holes, and placed in a container; thus, the position z spikes in the red region ('place' stage). Interestingly, despite the differing scale of force feedback between the two platforms, the magnitude of external force observed over time for the bolt removal task is similar for both platforms. This suggests that the realism and scaling of the force feedback is less important for accomplishing precise grasping tasks for disassembly than the directional guidance provided by the force feedback during contact.

Some tasks involved placing components after disassembly into a waste/recycle bin, like in the cover removal task, where most of the time was spent on post-grasp manipulation, i.e. the 'place' stage. Operators spent on average 39, and 37 seconds placing the objects which averaged 56\%, and 35\% of the task completion time with cobot and haptic platform, respectively. Around 30\% of the time was spent in the 'place' stage for the cobot platform; due to the lack of Cartesian control, the operator reached joint configurations where large reconfigurations were required between tasks to access the workspace. This is evident in Figure \ref{fig:Results-Rbolts-PosForce} around 150s. Comparing trial 4 in Figure \ref{fig:Results-Rcover-PosForce}, for which a smaller difference in task time was observed (SMD 0.85) suggests that coarse alignment with the correct grasp point is achieved on a similar timescale of $\sim$20s. For the haptic, a greater proportion of time is spent on fine alignment with the correct grasp point, while for the Franka, this time is predominantly spent on post-grasp manipulation.

For the module sorting task, significant improvements in the task completion time were observed, with an SMD of 2.31. Similarly to other tasks, improvements in coarse and fine alignment speed are observed, although the reduction in completion time is driven by time reductions over all stages of the task, including the `action' stage. Examining in Figure \ref{fig:Results-Suction-PosForce} shows that after positional alignment, an increase in the force (up to $\sim$25N) occurs indicating the attempt to reach a successful suction grip at the module surface. As with the previous tasks, the magnitude of force feedback is largely consistent between both platforms; however, for the action phase, the operator exhibited a more slow and conservative approach with the haptic device, characterised by a slow ramp in the normal (+Z) force. Intuitively, the operator uses a mix of visual and tactile cues to identify engagement of the suction cups with the surface. This implies a combination of factors can be attributed to this behaviour; in the first instance, from similar factors contributing to faster positional alignment with the Franka, such as 1:1 position mapping, while in the second instance, the operator can identify the correct force threshold more intuitively, as with direct manipulation of the slave arm, through 1:1 force mapping. Next, the `action' stage is followed by the 'place' stage (red region) where the module is removed from the scene. Contrasting the cover removal task, a much shorter period of time is spent both proportionally and in absolute terms on post-grasp manipulation with the Franka. Comparison of the trial breakdown in Figure \ref{fig:Results-Rcover-PosForce} and Figure \ref{fig:Results-Suction-PosForce} for these stages implies an effort-related component to the task performance, due to the higher load exerted by the cover (notable in the Z direction). Furthermore, the cover exhibits a more dispersed mass distribution in contrast with the module, where a majority of the weight is concentrated about the grasp point. Hence, positioning of the cover imposes a greater demand due to higher torque required to rotate the joints, which is largely abstracted for the Cartesian controlled haptic-Franka platform. This is corroborated by the more gradual and tortuous path seen in Figure \ref{fig:Results-Rcover-PosForce}. Therefore, realism of force feedback may be considered beneficial for the contact-rich stages of the task, but a telemanipulation framework incorporating 1:1 feedback of external torque should be aware of detrimental effects on manipulations in free space.

\begin{figure}
    \centering
    \begin{minipage}[b]{0.49\textwidth}
        \centering
        \includegraphics[width=\linewidth]{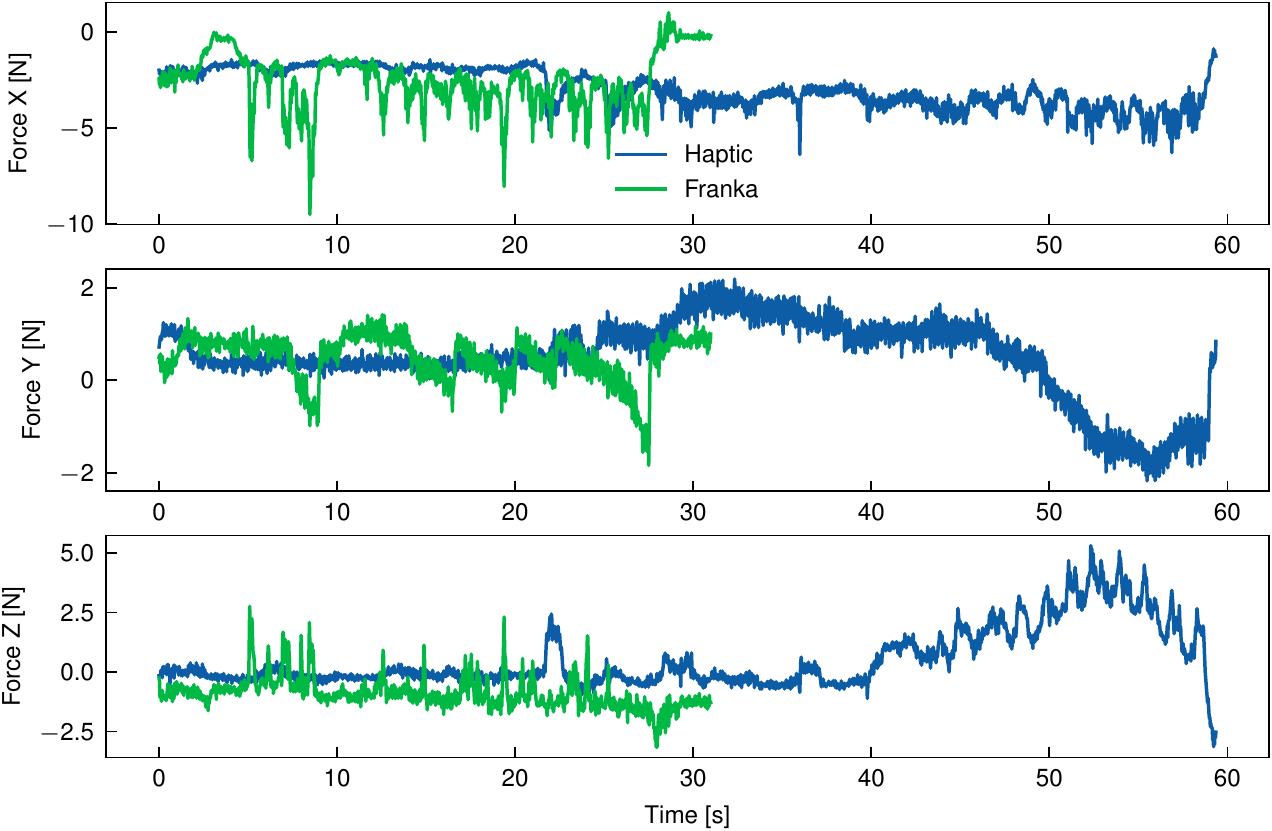}
        (a) Cutting forces for trial 3
    \end{minipage}\hspace{0.01\textwidth}
    \begin{minipage}[b]{0.4\textwidth}
        \centering
        \includegraphics[width=\linewidth]{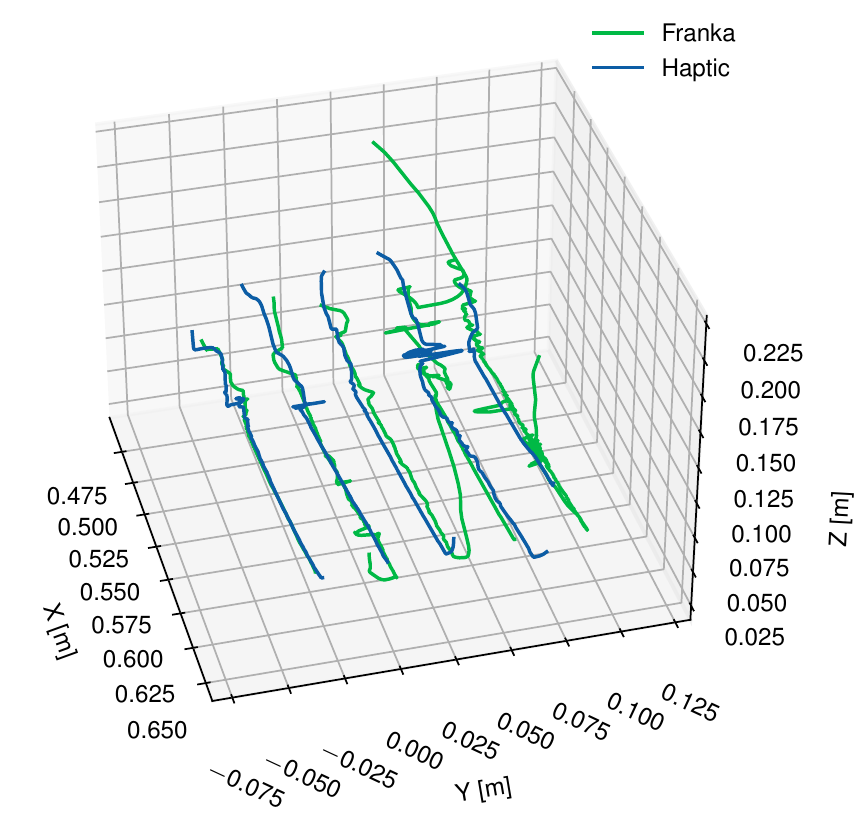}
        (b) Cutting paths, all trials
    \end{minipage}
    \caption{Comparison of cutting forces between each telerobotics platform for single trial, and overall paths adopted by each operator for all trials.}
    \label{fig:Results-Cutting-Graphs}
\end{figure}

For cutting, a small SMD of 0.41 suggests the effect on task completion time for the Franka platform is marginal relative to the trial-to-trial variation. Cutting is a complex task dependent on a wide range of material and tool-specific parameters, and a wide variation in completion times is observed between task trials. For the Franka, a greater proportion of time is spent performing the cutting task proper. This is on average achieved faster with the haptic device, however, a much greater proportion of time (55\%) is required for initial alignment of the tool with the desired path. The fine alignment stage in particular is unique for the cutting task in that the operator must rely purely on visual alignment with the desired path; this further suggests the reduction in completion time for the presented tasks is in large part related to the expanded workspace and 1:1 \emph{positional} mapping afforded with the Franka, rather than improvements in realism of the tactile feedback. However, during the cutting task, the realism of feedback is comparatively more important. This can be examined in Figure \ref{fig:Results-Cutting-Graphs}a, showing the force profile for the contact cutting task for both platforms. Notably, the force for the haptic trial steadily increases towards the end of the cutting operation, which can be attributed to drift of the operator setpoint away from the path in the transverse (Y) direction and an increase in the depth of cut (-Z). A similar effect can be observed in the transverse direction for the Franka, however, these are rapidly corrected during the course of the cutting task. This force feedback provides important cues to the operator that were otherwise ignored, or insufficient to prompt a response in the haptic case; similar force profiles were observed over the majority of the haptic trials.

The increased feedback capability of a setup with identical cobots has some notable disadvantages. From Figure \ref{fig:Results-Cutting-Graphs}b, it is clear that the cutting path adopted by the user in the Franka case study is subject to small variations along the length of the desired path. This is posited to be due to the lack of task space control, the larger motions achievable with the Franka, and low stiffness of the human arm. Owing to these factors, the master arm responds readily to the force feedback created by the executed motions during the cutting process, causing small deflections from the operator's desired path. This can be observed from the comparatively more variable and discontinuous force profile opposing the feed direction (+X) in Figure \ref{fig:Results-Cutting-Graphs}a. For higher strength materials, this effect is expected to increase in significance due to an increase in the feed-rate dependent cutting forces. In the absence of remediating strategies, this consequently reduces the quality of cut and increases physical demand on the user. However, remediating strategies such as scaling the force feedback provided to the user, or applying a Cartesian impedance behaviour to the master arm to guide the user along the path would have the effect of distorting the feedback, thus reducing the realism of the interaction.

For the module stack disassembly case study comprising 4 modules, total disassembly times averaged 25.8 minutes and 14.7 minutes for the haptic and Franka platforms respectively, representing a 43\% reduction in overall completion time when using two identical cobots. Considering module cover separation consisting of 4 linear cutting operations, it is estimated to require 8 minutes per module to expose the battery cells, reducing by 25\% for the identical cobot case. Overall disassembly times demonstrate that while disassembly via telemanipulation is feasible, several efficiency improvements and assistive strategies are required to improve disassembly speed. However, there is large variation in disassembly time estimates across battery models and techno-economic analyses in literature. For example, an estimate of 12 seconds for manual unscrewing and 6 seconds for manual removal of free components across multiple battery designs \citep{LANDER2023120437} implies large reductions in efficiency versus manual disassembly. Another technoeconomic analysis by \cite{ALFAROALGABA2020104461} for the Audi e-Tron hybrid estimates a total disassembly time of 30 minutes for pack-to-module disassembly. However, in \cite{batteries7040074}, extraction of the battery modules from the Smart ForFour was estimated at 10 minutes per module, and 135 minutes for disassembly of individual modules with 2 human personnel. Beyond module level, disassembly is hampered significantly by expensive cutting operations required to expose the individual battery cells. Broadly, this is corroborated across battery designs, chiefly due to the high level of design compartmentalisation and use of welds, glue and interference fit fasteners that are difficult to remove non-destructively \citep{LANDER2023120437, batteries7040074}.

\section{Conclusion} \label{sec:Conclusion}

This study demonstrates the telerobotic disassembly of a stack of modules from the Nissan Leaf 2011 battery pack. A comparative study between a master-slave setup consisting of a haptic device paired with a cobot, and two identical cobots examined the success rate and completion time for accomplishing unbolting, extraction of bolts, grasping and removal of the cover plate, sorting of modules with a suction gripper and contact cutting. While overall success rate was higher with the haptic device, a setup with identical paired cobots was found to significantly improve task completion times and time consistency across the entire set of disassembly tasks. This suggests that quality and realism of force feedback is comparatively less important for accomplishing precise manipulation tasks, but instead the 1:1 position mapping between master interface and slave and expanded workspace were main contributors to the efficiency improvements. While 1:1 mapping of torques between identical cobots was beneficial for grasping with a vacuum gripper and cutting, chiefly due to the enhanced realism of the interaction, providing strengthened tactile cues to the operator, this further had detrimental effects on operator physical effort and quality of cut.

Limitations of the present work are that only objective measures of task performance are considered under the assumption of familiarity with the necessary tasks and teleoperation platforms (expert operators). However, to obtain a holistic comparison it is necessary to consider operators of different experience levels, as well as consideration of subjective performance measures, such those established in NASA TLX that establish measures of cognitive load and effort for the sequence of disassembly tasks. Furthermore, although this study considers a disassembly case study of a Nissan Leaf 2011 battery pack, a wide range of EV batteries are in circulation whose designs differ broadly between manufacturers and models. Therefore, this study could be extended to consider a wider range of battery designs to achieve a more general overview of expected level of performance for the EV battery disassembly process.

\section*{Conflict of Interest Statement}
The authors declare that the research was conducted in the absence of any commercial or financial relationships that could be construed as a potential conflict of interest.

\section*{Author Contributions}
Conceptualization, A.R.; methodology, J.H., A.S. and A.R.; software, A.S., J.H. and A.A.; investigation, A.S., J.H., A.A. and A.R.; validation, J.H, A.S and A.R; data curation, J.H. and A.S.; formal analysis, J.H. and A.S.; visualisation, J.H. and A.S.; writing---original draft, C.A., J.H, A.S and A.R; writing---review \& editing, A.R., C.A. and J.H; resources, A.R. and R.S.; supervision A.R. and R.S.; funding acquisition, A.R. and R.S.; project administration, A.R.

\section*{Funding}
This research was conducted as part of the project called  ``Reuse and Recycling of Lithium-Ion Batteries'' (RELiB). This work was supported by the Faraday Institution [grant number FIRG005].

\section*{Acknowledgments}
The authors would like to acknowledge Christopher Gell for his technical support.

\bibliographystyle{Frontiers-Harvard} 

\bibliography{main.bib}

\end{document}